\documentclass{IEEEtran} 
\pdfoutput=1
\usepackage{tikz}
\usepackage{pgfplots}
\usepackage{aircraftshapes}
\usetikzlibrary{angles,quotes,shapes,arrows}
\usepackage{verbatim}
\pgfplotsset{compat=newest}
\pgfplotsset{every axis legend/.append style={%
		cells={anchor=west}}
}
\pgfplotsset{every y tick label/.append style={font=\footnotesize}}
\pgfplotsset{every x tick label/.append style={font=\footnotesize}}
\pgfplotsset{every axis x label/.append style={font=\footnotesize}}
\pgfplotsset{every axis y label/.append style={font=\footnotesize}}
\pgfplotsset{every axis legend/.append style={font=\footnotesize}}
\pgfplotsset{every axis title/.append style={font=\footnotesize}}
\usepgfplotslibrary{polar}
\tikzset{>=stealth'}
\usepgfplotslibrary{groupplots}
\usepackage{graphicx}
\DeclareGraphicsExtensions{.png}
\usepackage{amsmath}
\usepackage[noend]{algorithmic}
\usepackage{algorithm}
\usepackage{array}
\usepackage{fixltx2e}
\usepackage{float}
\usepackage{url}
\usepackage{booktabs}
\usepackage[capitalize]{cleveref}
\usepackage[per-mode=symbol,detect-all,binary-units=true]{siunitx}
\newcommand{\argmax}{\operatornamewithlimits{arg\,max}}
\usepackage{changepage}

\makeatletter
\def\blx@maxline{77}
\makeatother

\newcommand{\edit}[1]{{#1}}

\title{
	Deep Neural Network Compression for Aircraft Collision Avoidance Systems
}

\author{Kyle D. Julian$^{1}$ and Mykel J. Kochenderfer$^{2}$ and Michael P. Owen$^{3}$
	\thanks{$^{1}$Kyle D. Julian is a Ph.D. candidate in the Department of Aeronautics and Astronautics,
		Stanford University, Stanford, CA, 94305
		{\tt\small kjulian3@stanford.edu}}%
	\thanks{$^{2}$Mykel J. Kochenderfer is an Assistant Professor in the Department of Aeronautics and Astronautics,
		Stanford University, Stanford, CA, 94305
		{\tt\small mykel@stanford.edu}}%
	\thanks{$^{3}$Michael P. Owen is a member of the Technical Staff at Lincoln Laboratory, Massachusetts Institute of Technology, Lexington, MA, 02421
		{\tt\small michael.owen@ll.mit.edu}}%
}

\begin{document}
	\maketitle
	\thispagestyle{empty}
	\pagestyle{empty}

\begin{abstract}
	One approach to designing decision making logic for an aircraft collision avoidance system frames the problem as a Markov decision process and optimizes the system using dynamic programming. The resulting collision avoidance strategy can be represented as a numeric table. This methodology has been used in the development of the \edit{Airborne Collision Avoidance System X (ACAS X)} family of collision avoidance systems for manned and unmanned aircraft, but the high dimensionality of the state space leads to very large tables. To improve storage efficiency, a deep neural network \edit{is used to approximate} the table. With the use of an asymmetric loss function and a gradient descent algorithm, the parameters for this network can be trained to provide accurate estimates of table values while preserving the relative preferences of the possible advisories for each state. By training multiple networks to represent subtables, the network also decreases the required runtime for computing the collision avoidance advisory. Simulation studies show that the network improves the safety and efficiency of the collision avoidance system. Because only the network parameters need to be stored, the required storage space is reduced by a factor of 1000, enabling the collision avoidance system to operate using current avionics systems.
\end{abstract}

\maketitle

\section{Introduction}
Decades of research have explored a variety of approaches to designing decision making logic for aircraft collision avoidance systems for both manned and unmanned aircraft \cite{Kuchar2000}. Recent work on formulating the problem of collision avoidance as a partially observable Markov decision process (POMDP) has led to the development of the \edit{Airborne Collision Avoidance System X (ACAS X)} family of collision avoidance systems \cite{Kochenderfer2015chapter,Kochenderfer2012lljournal,Kochenderfer2011atc371}. The version for manned aircraft, ACAS Xa, is expected to become the next international standard for large commercial transport and cargo aircraft. The variant for unmanned aircraft, ACAS Xu, uses dynamic programming to determine horizontal or vertical resolution advisories in order to avoid collisions while minimizing disruptive alerts. ACAS Xu was successfully flight tested in 2014 using  NASA's Ikhana aircraft \cite{ACAS-XuTests}.

The dynamic programming process for creating the ACAS Xu horizontal decision making logic results in a large numeric lookup table that contains scores associated with different maneuvers from millions of different discrete states. The table is extremely large, requiring hundreds of gigabytes of floating point storage. A simple technique to reduce the size of the score table is to downsample the table after dynamic programming. To minimize the degradation in decision quality, states are removed in areas where the variation between values in the table are smooth. The downsampling reduces the size of the table by a factor of 180 from that produced by dynamic programming. For the rest of this paper, the downsampled ACAS Xu horizontal table \edit{is referred to} as the baseline, original table.

Even after downsampling, the current table requires over 2GB of floating point storage, too large for certified avionics systems \edit{\cite{youn2014software}. Although modern hardware can handle 2GB of storage, the certification process for aircraft computer hardware is expensive and time-consuming, so a solution capable of running on legacy hardware is desired \cite{AvionicsOpinion}. While there is no formal limit for floating point storage on legacy avionics, a representation occupying less than 120MB would be sufficient}. 

For an earlier version of ACAS Xa, block compression was introduced to take advantage of the fact that, for many discrete states, the scores for the available actions are identical \cite{Kochenderfer2013a}. One critical contribution of that work was the observation that the table could be stored in IEEE half-precision with no appreciable loss of performance. Block compression was adequate for the ACAS Xa tables that limit advisories to vertical maneuvers, but the ACAS Xu tables for horizontal maneuvers are much larger. Recent work explored a new algorithm that exploits the score table's natural symmetry to remove redundancy within the table \cite{julian2016policy}. However, results showed that this compression algorithm could not achieve sufficient reduction in storage before compromising performance.

Discretized score tables like this can be represented as Gaussian processes \cite{Engel2005} or kd-trees \cite{Munos2002}. \edit{Decision trees offer a way to compress the table by organizing the data into a tree structure to remove table redundancy. In addition a decision tree can increase compression by simplifying areas of the table with low variance, although this will result in a lossy compression. Decision trees are a popular machine learning algorithm and have been applied to numerous problems including land cover classification and energy consumption prediction \cite{pal2003assessment,tso2007predicting}.}

\edit{Other approaches to compressing the table seek to find a robust nonlinear function approximation that represents the table. Linear regression is popular for smaller datasets, but this approach does not generalize well for large datasets with many more examples than features. Support Vector Machines (SVM) are also a popular regression algorithm. By storing only the supporting vectors found by the algorithm, less data would need to be stored, effectively compressing the dataset. However, SVMs struggle with large datasets. Solving the SVM quadratic program has computational cost that scales at least with the square of the number of examples \cite{bordes2005fast}. Because the ACAS Xu table has millions of entries, SVMs would not be effective in regressing the score table.}

Neural networks, which can serve as a robust global function approximator when trained using supervised learning, can represent large datasets and are trained efficiently. \edit{Neural networks have been employed for regression applications in the aerospace field since the 1990s, including aircraft control systems, aircraft design, and impact detection on structural panels \cite{kim1997nonlinear,norgaard1997neural,worden2007application}. These applications use datasets with only a few hundred training examples, so small neural networks with one or two hidden layers are effective. However, to accurately represent 2GB of floating point values, a larger network is required.}

\edit{Recent works have shown that deep neural networks, or neural networks with more than two hidden layers, represent data more efficiently than shallow networks. It can be shown that the number of linear regions represented by neural networks grows exponentially with the depth and polynomially with the layer size, so a deeper network can represent more information than a shallow network \cite{montufar2014number}. In addition, it was shown that a three-layer network cannot be approximated by a two-layer to arbitrary accuracy unless the layer size is expanded exponentially \cite{eldan2016power}. Previously, neural networks were limited in depth due to activation saturation problems when using sigmoid activation functions \cite{glorot2010understanding}, but a new piecewise-linear activation function, rectified linear units (ReLUs), were shown to be well suited for training deep neural networks \cite{glorot2011deep}. These advancements enable neural networks to efficiently regress large amounts of data like the ACAS Xu collision avoidance table.}

This paper explores the use of deep neural networks for compressing the score table without loss of performance as measured by a set of safety and operational metrics. Using an asymmetric loss function during training ensures that the approximation is able to maintain the actions recommended by the original table while providing good estimates of the original score table values, but the required runtime is significantly increased. Methods for eliminating the runtime increase are explored that lead to an approach that produces advisories more quickly than the original table. Further studies into fine-tuning the network training process enable the neural network to predict table values with even greater accuracy.

Standard safety and operational performance metrics were used to evaluate the network performance in simulation. These metrics are calculated by simulating millions of encounters with varied encounter geometries and sensor noise. Although the deep neural network reduces the required memory by a factor of 1000, it also improves the performance of the ACAS Xu system on most performance metrics evaluated in this paper with only one operational metric slightly degraded. The neural network representation is a continuous function that smooths out artifacts from interpolation of the original discrete representation, allowing the network to surpass the system it was trained to represent.

Section \ref{sec:ScoreTable} provides an overview of the score table used in the ACAS Xu horizontal logic. Section \ref{sec:Comparisons} explores common machine learning approaches. Section \ref{sec:NN} describes the deep neural network approach to table compression. Section \ref{sec:ImproveRuntime} discusses ways to ensure online computation does not increase required runtime. Section \ref{sec:SmoothData} discusses techniques for making the table easier for the neural network to regress. Section \ref{sec:Results} discusses performance results computed in simulation. Conclusions are presented in Section \ref{sec:Conclusion}.

\section{Score Table \label{sec:ScoreTable}}
The ACAS Xu score table associates values to combinations of actions and state variables. The actions in the score table are the horizontal resolution advisories given to the ownship. These advisories tell the vehicle either it is Clear-of-Conflict (COC) or that it should turn left or right at one of two specified heading rates, \edit{\SI{1.5}{deg\per\second}} (weak) or \edit{\SI{3.0}{deg\per\second}} (strong). Hence, there are five possible actions: COC, weak left (WL), weak right (WR), strong left (SL), and strong right (SR). 

 There are seven state variables that define an aircraft encounter \cite{Kochenderfer2011atc371}:
\begin{enumerate}
	\item $\rho$ (ft): Distance from ownship to intruder
	\item $\theta$ (rad): Angle to intruder relative to ownship heading direction
	\item $\psi$ (rad): Heading angle of intruder relative to ownship heading direction
	\item $v_\text{own}$ (ft/s): Speed of ownship
	\item $v_\text{int}$ (ft/s): Speed of intruder
	\item $\tau$ (sec): Time until loss of vertical separation
	\item $a_\text{prev}$ (\edit{\si{deg\per\second}}): Previous advisory
\end{enumerate}

\begin{figure}[h]
	\centering
	
\def\layersep{1.6cm}
\def\layerseP{1.12cm}

\begin{tikzpicture}[scale=0.7]

\node[aircraft top,fill=black,minimum width=1cm,rotate=90,scale = 0.85] (Own) at (0,0) {} node [below,yshift=-0.5cm,font=\footnotesize] {Ownship};
\coordinate[label=right:$v_\text{own}$] (S0) at (0,2.5);

\node[aircraft top,fill=black,minimum width=1cm,rotate=166, scale = 0.85] (Int) at (5,3) {};
\node at (5,3) [above,xshift=1.1cm,font=\footnotesize] {Intruder};
\coordinate[label=above:$$] (IntN) at (5,5.5);
\coordinate[label=below:$v_\text{int}$] (S1) at (2.6,3.6);

\coordinate[label=below:$\rho$] (R) at (3.3,1.5);

\draw [thick, ->] (Own) -- (S0);
\draw [dashed,->] (Int) -- (IntN);
\draw [thick, ->] (Int) -- (S1);
            
\draw[dashed] (Own) -- (Int);

\pic [draw,-stealth,black,thick,dashed,angle radius=1.1cm,"$\psi$"{anchor=west,text = black, below}, angle eccentricity=1] {angle = IntN--Int--S1};
\pic [draw,-stealth,black,thick,dashed,angle radius=1.1cm,"$\theta$"{anchor=west,text = black, below}, angle eccentricity=1] {angle = S0--Own--Int};

\end{tikzpicture}
	\caption{Geometry for ACAS Xu \edit{h}orizontal \edit{l}ogic \edit{t}able}
	\label{geo_horizontal}
\end{figure}
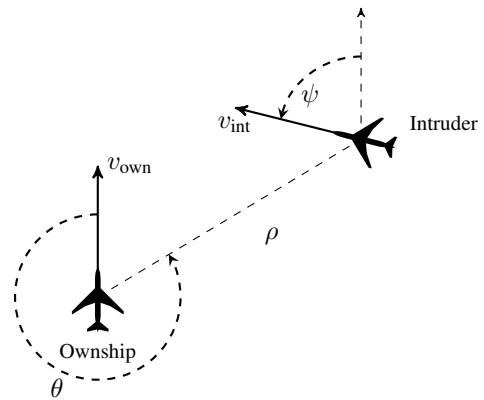

The first five state variables describe the geometry of a 2D encounter between two aircraft, as seen in \cref{geo_horizontal}. In addition, $\tau$ describes the encounter geometry vertically \edit{and extends the encounter geometries to 3D}. Lastly, $a_\text{prev}$ specifies the previous advisory to enable consistency when choosing the next advisory. The state variables are discretized, forming a seven dimensional grid with 120 million discrete points.

The optimal advisory for a given state can be extracted from the score table. If $Q$ is the real-valued function associated with the eight-dimensional score table, then the optimal action is
\begin{equation} \label{eq_optimal}
a^* = \argmax_a Q(\rho, \theta, \psi, v_\text{own}, v_\text{int}, \tau, a_\text{prev}, a)
\end{equation}
Furthermore, \Cref{eq_optimal} defines the score table's policy by mapping all possible states to actions.
In general, the values of the state variables, as determined by the sensors in real time, do not fall exactly on the grid points, in which case nearest-neighbor interpolation is used.

Since the surveillance sensors used by the aircraft are imperfect, there may be uncertainty in the current state measurement. To improve the robustness of the system to this uncertainty \cite{Chryssanthacopoulos2011jgcd}, ACAS Xu uses an unscented Kalman filter \cite{Julier2004} to arrive at a set of weighted state-space samples. These weighted samples can then be used to compute the best action:
\begin{equation}
\label{qmdp}
a^* = \argmax_a \sum_i b(s^{(i)})Q(s^{(i)}, a)
\end{equation}
where $s^{(i)}$ is the $i$th state sample and $b(s^{(i)})$ is its associated weight.

\edit{Although the state space specifies an encounter with a single intruder, multi-intruder encounters can use the same collision avoidance table by using utility fusion \cite{Kochenderfer2011atc371}. The action scores can be computed for each intruder and then fused into a single action by considering only the worst-case intruder or by summing the action scores for each intruder and taking the action with the best score \cite{Kochenderfer2011atc371}. Therefore, representing the collision avoidance policy as action scores allows the single intruder policy to be extended to multi-intruder scenarios.}

\section{Baseline Table Regression Methods \label{sec:Comparisons}}

With 600 million floating point numbers, the table requires over 2GB of storage. Compression algorithms that exploit table symmetries to reduce redundancies could not adequately compress the large score table \cite{Kochenderfer2013a,julian2016policy}. As a result, machine learning algorithms for lossy compression \edit{were explored}. 

\edit{Linear regression works well for small datasets, but with 120 million training examples and only seven features, a linear mapping from features to output values will be too simple to be accurate. Linear regression approximates the table values as $\hat{y}=XW$, where the weights $W$ can be optimized to minimize the squared error using $W=(X^TX)^{-1}X^Ty$. After calculating the weights, the root mean squared error (RMSE) is 17.9, which is approximately the standard deviation of the table values. A more complex model is necessary for this application.}

\edit{Although Support Vector Machines (SVM) can represent complex data, SVMs regress very slowly for large datasets, rendering them a poor choice for this application with a large number of training examples \cite{bordes2005fast}. Image compression algorithms like JPEG2000 or Huffman encoding can enable efficient storage and transmission of data files, but the data will have to be uncompressed and loaded into memory at runtime. Furthermore, the entire dataset must be loaded into memory to enable multi-intruder tracking, so the compressed representation must efficiently represent the entire table.}

\edit{With these considerations, decision trees are the best alternative to a neural network approach}. Decision trees are a popular regression algorithm and work well given large data sets. The size of the decision trees can be controlled by setting a maximum depth of the tree. At each level in the decision tree, an input variable is compared with a threshold, splitting the table data into two different sets. These decision nodes are chosen to minimize the variance in the split datasets through use of the Gini impurity \edit{\cite{lerman1984note}}. The algorithm begins by determining the root decision and then builds the tree layer by layer until the maximum depth is reached. The final layer is composed of leaf nodes, which store the average scores of the data in the leaf node. When making a prediction, a new state is passed through the decision tree, splitting at each decision node until arriving at a leaf node where the stored values become the predicted values. 

The amount of storage required to represent the decision tree can be computed by multiplying the number of nodes in the tree with the size of each node. Decision nodes will need to store the decision feature and threshold as well as pointers to children nodes, while leaf nodes will need to store the estimated table values. While larger trees can represent the score table more accurately, they require more storage. \edit{The Scikit-learn machine learning library was used to create decision trees for the ACAS Xu score table \cite{pedregosa2011scikit}.}

\begin{figure*}[h]
	\centering
	\begin{tikzpicture}[]
\begin{groupplot}[group style={horizontal sep=0.7cm, group size=4 by 1}, width= 5.0cm, height = 4.3cm]
\nextgroupplot [title={\textbf{(a)}},every axis title/.style={above,at={(0,1)}},xlabel = {Downrange (kft)}, ylabel = {Crossrange (kft)}, enlargelimits = false, axis on top]\addplot [point meta min=-3, point meta max=4.5] graphics [xmin=-5.5, xmax=25.5, ymin=-5.5, ymax=25.5] {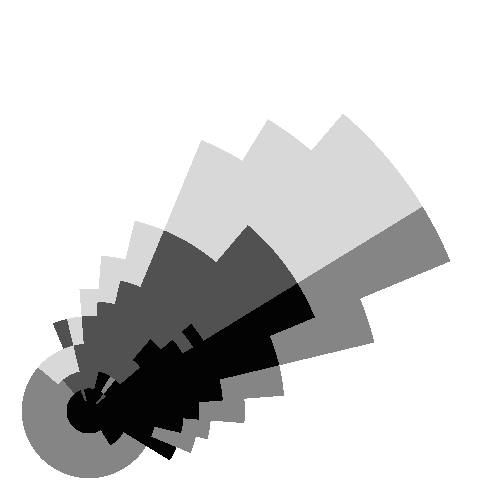};
\node[aircraft top,draw=white,fill=black,minimum width=1cm,rotate=0,scale = 0.55] at (axis cs:0.0, 0.0) {};
\node[aircraft top,draw=white,fill=black,minimum width=1cm,rotate=-90,scale = 0.55] at (axis cs:20,20) {};

\node[text=black] at (20,1)  {\scriptsize COC};
\node[text=white] at (6,2)  {\scriptsize SL};
\node[text=white] at (6,7.5)  {\scriptsize SR};
\node[text=white] at (15.5,5.5)  {\scriptsize WL};
\node[text=black] at (15.5,13.5)  {\scriptsize WR};

\nextgroupplot [title={\textbf{(b)}},every axis title/.style={above,at={(0,1)}},xlabel = {Downrange (kft)}, enlargelimits = false, axis on top]\addplot [point meta min=-3, point meta max=4.5] graphics [xmin=-5.5, xmax=25.5, ymin=-5.5, ymax=25.5] {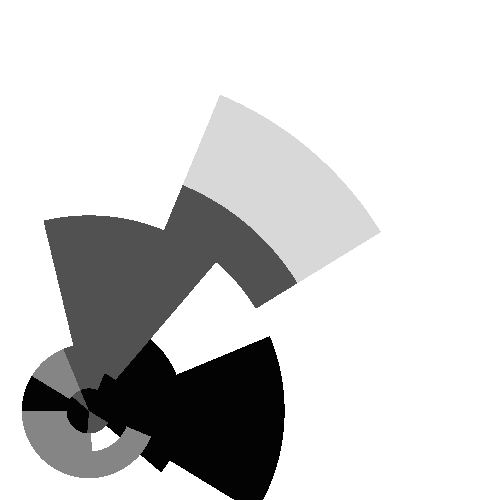};
\node[aircraft top,draw=white,fill=black,minimum width=1cm,rotate=0,scale = 0.55] at (axis cs:0.0, 0.0) {};
\node[aircraft top,draw=white,fill=black,minimum width=1cm,rotate=-90,scale = 0.55] at (axis cs:20,20) {};

\nextgroupplot [title={\textbf{(c)}},every axis title/.style={above,at={(0,1)}},xlabel = {Downrange (kft)},  enlargelimits = false, axis on top]\addplot [point meta min=-3, point meta max=4.5] graphics [xmin=-5.5, xmax=25.5, ymin=-5.5, ymax=25.5] {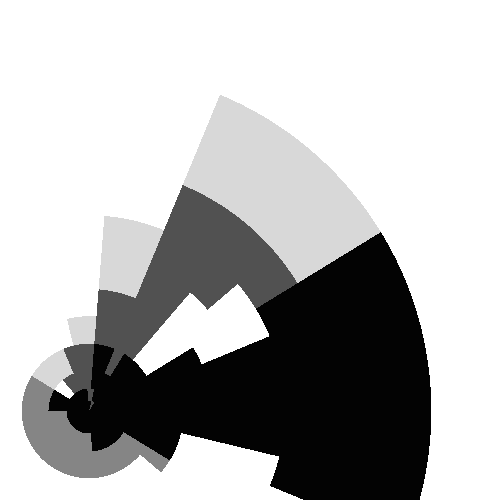};
\node[aircraft top,draw=white,fill=black,minimum width=1cm,rotate=0,scale = 0.55] at (axis cs:0.0, 0.0) {};
\node[aircraft top,draw=white,fill=black,minimum width=1cm,rotate=-90,scale = 0.55] at (axis cs:20,20) {};

\nextgroupplot [title={\textbf{(d)}},every axis title/.style={above,at={(0,1)}}, xlabel = {Downrange (kft)},  enlargelimits = false, axis on top]\addplot [point meta min=-3, point meta max=4.5] graphics [xmin=-5.5, xmax=25.5, ymin=-5.5, ymax=25.5] {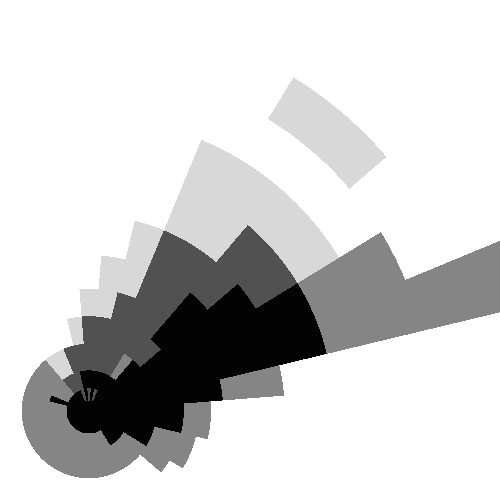};
\node[aircraft top,draw=white,fill=black,minimum width=1cm,rotate=0,scale = 0.55] at (axis cs:0.0, 0.0) {};
\node[aircraft top,draw=white,fill=black,minimum width=1cm,rotate=-90,scale = 0.55] at (axis cs:20,20) {};
\end{groupplot}

\end{tikzpicture}
	\caption{\edit{Policies for (a) original table, (b) 2.56MB decision tree, (c) 19.4MB decision tree, and (d) 126MB decision tree}}
	\label{fig_DT}
\end{figure*}
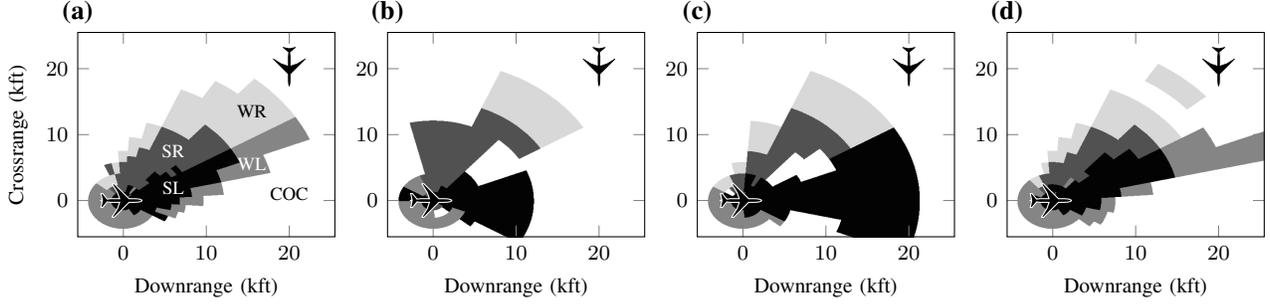

One way to assess the quality of the compression is to plot the actions recommended by the original table and the compressed strategies. Because the input space is seven dimensional, the plots only show variations in $\rho$ and $\theta$, which are converted to Cartesian coordinates, while the other inputs are constant. \Cref{fig_DT} shows top-down views of encounters with the ownship centered at the origin and flying in the direction indicated while the intruder vehicle is flying in the direction shown by the aircraft in the upper right corner of the plots. The color at each point in the plot shows the advisory the collision avoidance system would issue if the intruder were at that location.
\Cref{fig_DT} shows the policy plots of regressed decision trees of different sizes. Increasing the maximum depth improves accuracy, although there are still errors between the decision tree and original policy.

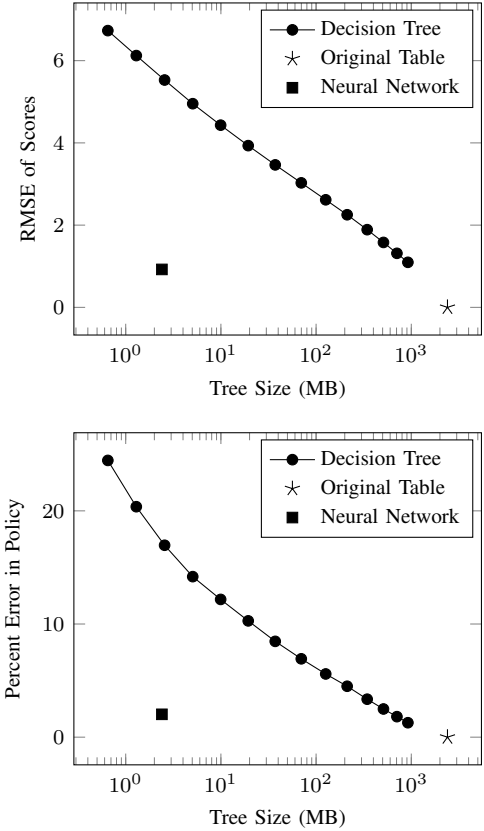
\begin{figure}[h]
	\centering
	\begin{tikzpicture}

\begin{groupplot}[group style={vertical sep = 1.3cm, group size=1 by 2},every x tick label/.append style={font=\footnotesize},every y tick label/.append style={font=\footnotesize}, width=7.0cm,height=6.0cm]
	
\nextgroupplot [ylabel = {RMSE of Scores}, xlabel = {Tree Size (MB)}, xmode = {log}]
\addplot+ [black, mark options={fill=black}]coordinates {
	(0.649060249, 6.73086302575)
	(1.29145145, 6.12349271756)
	(2.56177425, 5.52929891583)
	(5.05920124, 4.95331389655)
	(9.93845844, 4.43121337203)
	(19.3654032, 3.93336944806)
	(37.2186956, 3.4650376225)
	(69.8796301, 3.02781713659)
	(126.247155, 2.61591040081)
	(212.0, 2.2521141241)
	(344.181556, 1.88916095078)
	(510.217332, 1.5790731209)
	(706.68534, 1.31416966798)
	(921.457932, 1.09533973342)
};
\addlegendentry{Decision Tree}
\addplot+ [black, mark options={fill=black, scale=1.5}, only marks, mark=star]coordinates {
	(2400, 0.0)
};
\addlegendentry{Original Table}

\addplot+ [black, mark options={fill=black}, only marks, mark=square*]coordinates {
	(2.4, 0.92271571847)
};
\addlegendentry{Neural Network}

\nextgroupplot [ylabel = {Percent Error in Policy}, xlabel = {Tree Size (MB)}, xmode = {log}]
\addplot+ [black, mark options={fill=black}]coordinates {
	(0.649060249, 24.458113000000004)
	(1.29145145, 20.366733999999997)
	(2.56177425, 16.960143000000006)
	(5.05920124, 14.186458000000002)
	(9.93845844, 12.173639999999997)
	(19.3654032, 10.277552)
	(37.2186956, 8.466903999999998)
	(69.8796301, 6.923984000000005)
	(126.247155, 5.586237000000005)
	(212.0, 4.5000000000000036)
	(344.181556, 3.3554879999999954)
	(510.217332, 2.4906479999999953)
	(706.68534, 1.8124229999999963)
	(921.457932, 1.2759940000000025)
};
\addlegendentry{Decision Tree}
\addplot+ [black, mark options={fill=black, scale=1.5}, only marks, mark=star]coordinates {
	(2400, 0.0)
};
\addlegendentry{Original Table}

\addplot+ [black, mark=square*, mark options={fill=black},only marks]coordinates {
	(2.4, 2.02)
};
\addlegendentry{Neural Network}
\end{groupplot}

\end{tikzpicture}
	\caption{Decision tree Pareto frontiers \edit{for accuracy in scores (top) and policy (bottom)}}
	\label{fig_DT_pareto}
\end{figure}

In addition, aggregate metrics were computed to assess compression quality. The decision trees can be used to predict the scores for every discretized state represented by original table. By comparing the predicted scores with the original table scores, the root mean squared error (RMSE) of the values can be computed. In addition, the overall policy error is calculated using \cref{eq_optimal} and comparing the optimal actions of the decision tree and original table. The tradeoff between tree size and compression error is plotted in \cref{fig_DT_pareto}. Using decision trees that are at most \SI{100}{\mega\byte} in size, the RMSE exceeds 3.0 with a policy error rate of over 6\%, which could result in significant changes to system performance. By contrast, the neural network representation, as discussed in the remainder of this paper, can accurately represent the table values with only \SI{2.4}{\mega\byte} of storage.

\section{Neural Network Compression \label{sec:NN}}
This section explains the development of a deep neural network approximation of the score table.

\subsection{Neural Network Formulation}
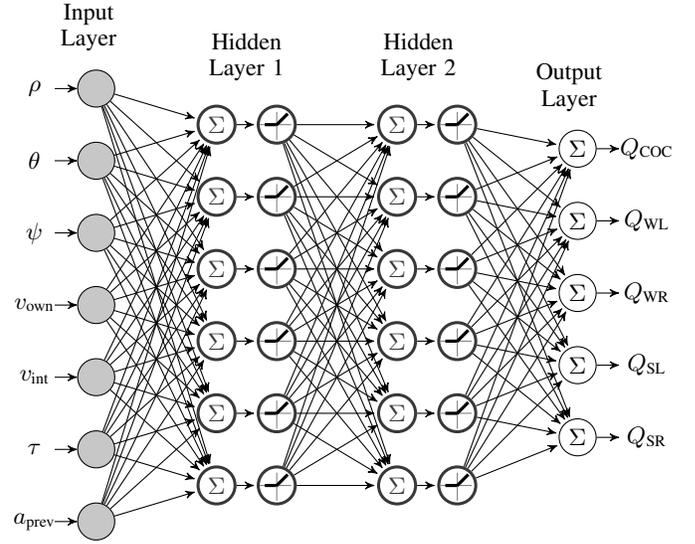
\begin{figure}
	\begin{adjustwidth*}{}{-1.4em}
		
\def\layerSep{2.0cm}
\def\reluSep{1.0cm}
\def\numIn{7}
\def\numHid{6}
\def\numOut{5}
\def\nodeSepY{1.2cm}
\def\labelSepY{0.9cm}
\def\annotDist{0.82cm}
\def\annotDistR{0.92cm}

\begin{tikzpicture}[shorten >=1pt,->,draw=black, node distance=\layerSep,scale = 0.8]
\tikzset{channel/.style={ 
		minimum size=8mm, minimum width=8mm,
		append after command={
			([shift={(16.0\pgflinewidth,0)}]\tikzlastnode.west)edge[-,thin,color=gray] ([shift={(-16.0\pgflinewidth,0)}]\tikzlastnode.east)
			([shift={(0,18*\pgflinewidth)}]\tikzlastnode.south)edge[-,thin,color=gray] ([shift={(0,-18*\pgflinewidth)}]\tikzlastnode.north)

			([shift={(+22.0\pgflinewidth,0)}]\tikzlastnode.west)edge[-,very thick, color=black] ([shift={(2.5\pgflinewidth,0)}]\tikzlastnode.center)
			
			([shift={(-0.5\pgflinewidth,-0.5\pgflinewidth)}]\tikzlastnode.center)edge[-, very thick, color=black]([shift={(-21\pgflinewidth,-21\pgflinewidth)}]\tikzlastnode.north east)				
		}
	}
}

\tikzstyle{every pin edge}=[<-,shorten >=5pt]
\tikzstyle{neuron}=[draw,circle,fill=black!25,minimum size=14pt,inner sep=0pt]
\tikzstyle{input neuron}=[neuron, fill=black!22!white];
\tikzstyle{output neuron}=[neuron, fill=white];
\tikzstyle{hidden neuron}=[neuron, color=white!22!black,fill=white, line width=0.4mm];
\tikzstyle{annot} = [text width=4em, text centered];
\tikzstyle{annot2} = [text width=4em, text centered];
\tikzstyle{relu}=[-,minimum size=5pt,fill=black];

\foreach \name / \y in {1,...,\numIn}
\node[input neuron, pin=left:] (I-\name) at (0,-\y*\nodeSepY) {};

\foreach \name / \y in {1,...,\numHid}{
	\path[yshift=-\nodeSepY/2]
	node[hidden neuron] (H1_1-\name) at (\layerSep,-\y*\nodeSepY) {\small $\Sigma$};
	\path[yshift=-\nodeSepY/2]
	node[hidden neuron] (H1_2-\name) at (\layerSep+\reluSep,-\y*\nodeSepY) {};
}

\foreach \name / \y in {1,...,\numHid}{
	\path[yshift=-\nodeSepY/2]
	node[hidden neuron] (H2_1-\name) at (\layerSep*2+\reluSep,-\y*\nodeSepY) {\small $\Sigma$};
	\path[yshift=-\nodeSepY/2]
	node[hidden neuron] (H2_2-\name) at (\layerSep*2+2*\reluSep,-\y*\nodeSepY) {};
}

\foreach \name / \y in {1,...,\numOut}{
	\path[yshift=-1.0cm]
	node[output neuron,pin={[pin edge={->}]right:}] (O-\name) at (\layerSep*3+2*\reluSep,-\y*\nodeSepY) {\small $\Sigma$};
} 
\foreach \source in {1,...,\numIn}
\foreach \dest in {1,...,\numHid}
\path (I-\source) edge (H1_1-\dest);

\foreach \source in {1,...,\numHid}
\path (H1_1-\source) edge (H1_2-\source);

\foreach \source in {1,...,\numHid}
\foreach \dest in {1,...,\numHid}
\path (H1_2-\source) edge (H2_1-\dest);
\foreach \source in {1,...,\numHid}
\path (H2_1-\source) edge (H2_2-\source);

\foreach \source in {1,...,\numHid}
\foreach \dest in {1,...,\numOut}
\path (H2_2-\source) edge (O-\dest);

\node[annot,above of=H1_1-1, node distance=\labelSepY, xshift=\reluSep/2.5, font=\small] (hiddenLabel) {\baselineskip=10pt Hidden Layer 1\par};
\node[annot,left of=hiddenLabel, node distance = \layerSep*1.05, yshift=\nodeSepY/3.0,font=\small] {\baselineskip=10pt Input Layer\par};
\node[annot,right of=hiddenLabel,node distance= \layerSep+\reluSep/3.5,font=\small](hiddenLabel2) {\baselineskip=10pt Hidden Layer 2\par};
\node[annot,right of=hiddenLabel2, node distance=\layerSep, yshift=-\nodeSepY/3.0,font=\small] {\baselineskip=10pt Output Layer\par};

\node[annot,left of=I-1, node distance=\annotDist,font=\small]{$\rho$};
\node[annot,left of=I-2, node distance=\annotDist,font=\small]{$\theta$};
\node[annot,left of=I-3, node distance=\annotDist,font=\small]{$\psi$};
\node[annot,left of=I-4, node distance=\annotDist,font=\small]{$v_\text{own}$};
\node[annot,left of=I-5, node distance=\annotDist,font=\small]{$v_\text{int}$};
\node[annot,left of=I-6, node distance=\annotDist,font=\small]{$\tau$};
\node[annot,left of=I-7, node distance=\annotDist,font=\small]{$a_\text{prev}$};

\node[annot,right of=O-1, node distance=\annotDistR,font=\small]{$Q_\text{COC}$};
\node[annot,right of=O-2, node distance=\annotDistR,font=\small]{$Q_\text{WL}$};
\node[annot,right of=O-3, node distance=\annotDistR,font=\small]{$Q_\text{WR}$};
\node[annot,right of=O-4, node distance=\annotDistR,font=\small]{$Q_\text{SL}$};
\node[annot,right of=O-5, node distance=\annotDistR,font=\small]{$Q_\text{SR}$};

\foreach \source in {1,...,\numHid}{
	\node[channel] at (H1_2-\source.center){};
	\node[channel] at (H2_2-\source.center){};
}
\end{tikzpicture}
	\end{adjustwidth*}
	\caption{Neural network diagram}
	\label{fig_network}
\end{figure}

Rather than storing all of the \edit{table} scores explicitly or using a tree approximation, the values of the table can be represented as a \edit{nonlinear}, parametric function that takes as input the values of the state variables and outputs the scores of the various actions. Instead of storing the table itself, only the parameters of the function need to be stored, which could significantly decrease the amount of storage required to represent the table. Deep neural networks are large, \edit{nonlinear} functions that can be trained to approximate complex multidimensional target data \cite{krizhevsky2012imagenet}. A feed-forward neural network is composed of inputs that are weighted and summed into a layer of perceptrons. The value at each perceptron then passes through an activation function before being weighted and summed again to form the next layer of perceptrons. In a deep neural network, there are multiple layers, called hidden layers, before reaching the last layer, or output layer, which represents the function approximation for the score table. The weights and biases of the network can be trained so that a given set of inputs will accurately compute the score table values. The choices of the neural network architecture and features can help the network to train quickly and accurately. \Cref{fig_network} shows an example diagram of a fully-connected network with two hidden layers and rectified linear unit activations (ReLU) \cite{ReLU}, which only allow positive inputs to pass through to the next layer.

The deep neural network uses fully-connected feed-forward layers with ReLU activation after each hidden layer. The network has seven inputs, one for each of the seven state variables. The output layer consists of five output nodes, one for each possible advisory\edit{: $Q_\text{COC}$, $Q_\text{WL}$, $Q_\text{WR}$, $Q_\text{SL}$, and $Q_\text{SR}$}. With this approach, one forward pass through the network computes the score values for each of the five advisories.

\subsection{Loss Function}
Initially, the network parameters are random, and the network performs poorly. A loss function is used to compute the network error, and the gradient of the loss is back-propagated through the network to update network parameters and improve performance. For typical regression problems, mean squared error (MSE) is used because it is simple, differentiable, and fast to compute. When applied to the problem of learning score table values, MSE gives accurate approximations. However, there is no longer a guarantee that the optimal advisory remains the same. For many of the states in the score table, the difference between the scores of the first and second best advisories is relatively small. When MSE fails to maintain the order of the actions, the network's collision avoidance strategy can be very different from that of the original table.

Predicting the optimal action given a set of inputs is a classification problem often solved with categorical cross entropy loss \cite{crossent}. This approach can predict optimal actions well, but there is no regard for representing the score values. It is important to capture the score values too in order to compute the optimal advisory over a weighted set of states using \cref{qmdp}. 

\begin{figure}[h]
	\centering
%
%
\begin{tikzpicture}

\begin{axis}[%
width=2.9in,
height=1.6in,
at={(0.758in,0.448in)},
scale only axis,
separate axis lines,
every outer x axis line/.append style={black},
every x tick label/.append style={font=\color{black},font=\footnotesize},
xmin=-1,
xmax=1,
xlabel={Score Estimate Error},
every outer y axis line/.append style={black},
every y tick label/.append style={font=\color{black},font=\footnotesize},
ymin=0,
ymax=20,
ylabel={Loss},
ylabel style={font=\footnotesize},
xlabel style={font=\footnotesize},
axis background/.style={fill=white},
every axis x label/.append style={font=\footnotesize},
every axis y label/.append style={font=\footnotesize},
legend style={at={(0.95,0.97)},anchor=north east,legend cell align=left,align=left,fill=white,font=\footnotesize}
]
\addplot [color=black,solid,line width=2.0pt]
  table[row sep=crcr]{%
1	1\\
0.97979797979798	0.960004081216202\\
0.95959595959596	0.92082440567289\\
0.939393939393939	0.882460973370064\\
0.919191919191919	0.844913784307724\\
0.898989898989899	0.808182838485869\\
0.878787878787879	0.7722681359045\\
0.858585858585859	0.737169676563616\\
0.838383838383838	0.702887460463218\\
0.818181818181818	0.669421487603306\\
0.797979797979798	0.636771757983879\\
0.777777777777778	0.604938271604938\\
0.757575757575758	0.573921028466483\\
0.737373737373737	0.543720028568513\\
0.717171717171717	0.514335271911029\\
0.696969696969697	0.485766758494031\\
0.676767676767677	0.458014488317519\\
0.656565656565657	0.431078461381492\\
0.636363636363636	0.40495867768595\\
0.616161616161616	0.379655137230895\\
0.595959595959596	0.355167840016325\\
0.575757575757576	0.331496786042241\\
0.555555555555556	0.308641975308642\\
0.535353535353535	0.286603407815529\\
0.515151515151515	0.265381083562902\\
0.494949494949495	0.24497500255076\\
0.474747474747475	0.225385164779104\\
0.454545454545455	0.206611570247934\\
0.434343434343434	0.188654218957249\\
0.414141414141414	0.17151311090705\\
0.393939393939394	0.155188246097337\\
0.373737373737374	0.139679624528109\\
0.353535353535353	0.124987246199367\\
0.333333333333333	0.111111111111111\\
0.313131313131313	0.0980512192633405\\
0.292929292929293	0.0858075706560555\\
0.272727272727273	0.0743801652892562\\
0.252525252525252	0.0637690031629425\\
0.232323232323232	0.0539740842771146\\
0.212121212121212	0.0449954086317723\\
0.191919191919192	0.0368329762269156\\
0.171717171717172	0.0294867870625446\\
0.151515151515151	0.0229568411386593\\
0.131313131313131	0.0172431384552597\\
0.111111111111111	0.0123456790123457\\
0.0909090909090909	0.00826446280991736\\
0.0707070707070707	0.0049994898479747\\
0.0505050505050505	0.0025507601265177\\
0.0303030303030303	0.000918273645546371\\
0.0101010101010101	0.000102030405060707\\
-0.0101010101010102	0.00204060810121419\\
-0.0303030303030303	0.0183654729109274\\
-0.0505050505050506	0.0510152025303542\\
-0.0707070707070707	0.099989796959494\\
-0.0909090909090908	0.165289256198347\\
-0.111111111111111	0.246913580246914\\
-0.131313131313131	0.344862769105193\\
-0.151515151515152	0.459136822773187\\
-0.171717171717172	0.589735741250893\\
-0.191919191919192	0.736659524538312\\
-0.212121212121212	0.899908172635446\\
-0.232323232323232	1.07948168554229\\
-0.252525252525253	1.27538006325885\\
-0.272727272727273	1.48760330578512\\
-0.292929292929293	1.71615141312111\\
-0.313131313131313	1.96102438526681\\
-0.333333333333333	2.22222222222222\\
-0.353535353535354	2.49974492398735\\
-0.373737373737374	2.79359249056219\\
-0.393939393939394	3.10376492194674\\
-0.414141414141414	3.43026221814101\\
-0.434343434343434	3.77308437914498\\
-0.454545454545455	4.13223140495868\\
-0.474747474747475	4.50770329558208\\
-0.494949494949495	4.8995000510152\\
-0.515151515151515	5.30762167125804\\
-0.535353535353535	5.73206815631058\\
-0.555555555555556	6.17283950617284\\
-0.575757575757576	6.62993572084481\\
-0.595959595959596	7.1033568003265\\
-0.616161616161616	7.5931027446179\\
-0.636363636363636	8.09917355371901\\
-0.656565656565657	8.62156922762983\\
-0.676767676767677	9.16028976635037\\
-0.696969696969697	9.71533516988062\\
-0.717171717171717	10.2867054382206\\
-0.737373737373737	10.8744005713703\\
-0.757575757575758	11.4784205693297\\
-0.777777777777778	12.0987654320988\\
-0.797979797979798	12.7354351596776\\
-0.818181818181818	13.3884297520661\\
-0.838383838383838	14.0577492092644\\
-0.858585858585859	14.7433935312723\\
-0.878787878787879	15.44536271809\\
-0.898989898989899	16.1636567697174\\
-0.919191919191919	16.8982756861545\\
-0.939393939393939	17.6492194674013\\
-0.95959595959596	18.4164881134578\\
-0.97979797979798	19.2000816243241\\
-1	20\\
};
\addlegendentry{Optimal Advisories};

\addplot [color=black,dashed,line width=2.0pt]
  table[row sep=crcr]{%
1	5\\
0.97979797979798	4.80002040608101\\
0.95959595959596	4.60412202836445\\
0.939393939393939	4.41230486685032\\
0.919191919191919	4.22456892153862\\
0.898989898989899	4.04091419242934\\
0.878787878787879	3.8613406795225\\
0.858585858585859	3.68584838281808\\
0.838383838383838	3.51443730231609\\
0.818181818181818	3.34710743801653\\
0.797979797979798	3.1838587899194\\
0.777777777777778	3.02469135802469\\
0.757575757575758	2.86960514233241\\
0.737373737373737	2.71860014284257\\
0.717171717171717	2.57167635955515\\
0.696969696969697	2.42883379247016\\
0.676767676767677	2.29007244158759\\
0.656565656565657	2.15539230690746\\
0.636363636363636	2.02479338842975\\
0.616161616161616	1.89827568615447\\
0.595959595959596	1.77583920008162\\
0.575757575757576	1.6574839302112\\
0.555555555555556	1.54320987654321\\
0.535353535353535	1.43301703907765\\
0.515151515151515	1.32690541781451\\
0.494949494949495	1.2248750127538\\
0.474747474747475	1.12692582389552\\
0.454545454545455	1.03305785123967\\
0.434343434343434	0.943271094786246\\
0.414141414141414	0.857565554535251\\
0.393939393939394	0.775941230486685\\
0.373737373737374	0.698398122640547\\
0.353535353535353	0.624936230996837\\
0.333333333333333	0.555555555555556\\
0.313131313131313	0.490256096316702\\
0.292929292929293	0.429037853280277\\
0.272727272727273	0.371900826446281\\
0.252525252525252	0.318845015814713\\
0.232323232323232	0.269870421385573\\
0.212121212121212	0.224977043158861\\
0.191919191919192	0.184164881134578\\
0.171717171717172	0.147433935312723\\
0.151515151515151	0.114784205693297\\
0.131313131313131	0.0862156922762983\\
0.111111111111111	0.0617283950617285\\
0.0909090909090909	0.0413223140495868\\
0.0707070707070707	0.0249974492398735\\
0.0505050505050505	0.0127538006325885\\
0.0303030303030303	0.00459136822773186\\
0.0101010101010101	0.000510152025303536\\
-0.0101010101010102	0.000102030405060709\\
-0.0303030303030303	0.000918273645546371\\
-0.0505050505050506	0.00255076012651771\\
-0.0707070707070707	0.0049994898479747\\
-0.0909090909090908	0.00826446280991734\\
-0.111111111111111	0.0123456790123457\\
-0.131313131313131	0.0172431384552597\\
-0.151515151515152	0.0229568411386593\\
-0.171717171717172	0.0294867870625446\\
-0.191919191919192	0.0368329762269156\\
-0.212121212121212	0.0449954086317723\\
-0.232323232323232	0.0539740842771146\\
-0.252525252525253	0.0637690031629426\\
-0.272727272727273	0.0743801652892562\\
-0.292929292929293	0.0858075706560556\\
-0.313131313131313	0.0980512192633405\\
-0.333333333333333	0.111111111111111\\
-0.353535353535354	0.124987246199367\\
-0.373737373737374	0.139679624528109\\
-0.393939393939394	0.155188246097337\\
-0.414141414141414	0.17151311090705\\
-0.434343434343434	0.188654218957249\\
-0.454545454545455	0.206611570247934\\
-0.474747474747475	0.225385164779104\\
-0.494949494949495	0.24497500255076\\
-0.515151515151515	0.265381083562902\\
-0.535353535353535	0.286603407815529\\
-0.555555555555556	0.308641975308642\\
-0.575757575757576	0.331496786042241\\
-0.595959595959596	0.355167840016325\\
-0.616161616161616	0.379655137230895\\
-0.636363636363636	0.404958677685951\\
-0.656565656565657	0.431078461381492\\
-0.676767676767677	0.458014488317519\\
-0.696969696969697	0.485766758494031\\
-0.717171717171717	0.514335271911029\\
-0.737373737373737	0.543720028568514\\
-0.757575757575758	0.573921028466483\\
-0.777777777777778	0.604938271604938\\
-0.797979797979798	0.636771757983879\\
-0.818181818181818	0.669421487603306\\
-0.838383838383838	0.702887460463218\\
-0.858585858585859	0.737169676563616\\
-0.878787878787879	0.7722681359045\\
-0.898989898989899	0.808182838485869\\
-0.919191919191919	0.844913784307724\\
-0.939393939393939	0.882460973370064\\
-0.95959595959596	0.92082440567289\\
-0.97979797979798	0.960004081216203\\
-1	1\\
};
\addlegendentry{Suboptimal Advisories};

\end{axis}
\end{tikzpicture}%
	\caption{Asymmetric loss function penalties}
	\label{fig_loss}
\end{figure}
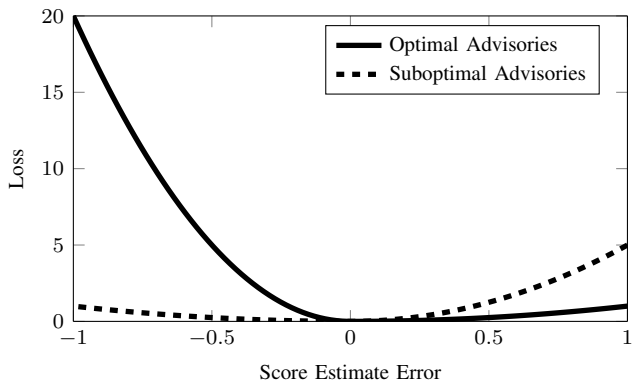

To get the numeric accuracy of MSE with the classification accuracy of categorical cross entropy, an asymmetric version of MSE was used. Asymmetric loss functions have been used to train neural networks when positive or negative errors are not identical \cite{Crone2002}. The asymmetric loss function for collision avoidance, shown in \cref{fig_loss} is based on MSE, but it increases the penalty by a factor when the neural network under-estimates the score of optimal advisories or over-estimates the score of sub-optimal advisories. The factor applied to the optimal advisory is four times greater than the suboptimal advisories to balance the fact that there are four sub-optimal advisories for every optimal advisory. Because the network parameters are updated to minimize the loss, the neural network attempts to eliminate any errors in the score predictions. If the network predictions have small errors, the asymmetric loss function encourages the network to over-estimate the scores of the optimal advisories while under-estimating the scores of suboptimal advisories, which will not change the policy when using \cref{eq_optimal}. Therefore, the asymmetric loss function encourages the neural network to maintain the optimal advisories of the score table while also learning accurate representations of the table values.

\begin{figure}[h]
	\centering
	\begin{tikzpicture}
	\input{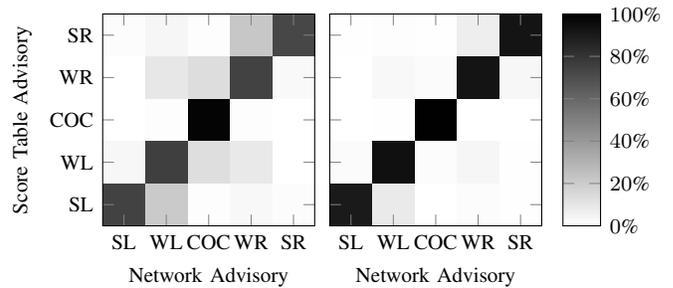}
	\end{tikzpicture}
	\caption{\edit{Confusion matrices for nominal (left) and asymmetric (right) MSE losses}}
	\label{fig_confusion}
\end{figure}

The advantage of the asymmetric loss function over regular MSE can be visualized through the confusion matrices shown in \cref{fig_confusion}. Each row is normalized to add up to 100\% and represents the percentage of states where the neural network selected a particular advisory given the advisory of the original table. 

The entries on the main diagonal represent the percentage of advisories correctly classified by the neural network, while the off-diagonal entries show mis-classifications. For the turning advisories, the asymmetric loss function is much better at maintaining advisories and achieves on-diagonal percentages of 90--94\% while the nominal MSE loss function maintains advisories only 72--74\% of the time.

\subsection{Model Architecture}
Optimizing the network architecture can be challenging because there are many parameters to vary and evaluation of the different architectures can be slow. One important consideration in training deep neural networks is to select and tune the optimizer used to update the parameters and weights of the network. Different optimizers were evaluated including RMSprop \cite{rmsprop}, Adagrad \cite{Adagrad}, Adadelta \cite{Adadelta}, and Adam \cite{Adam}. A variant of Adam, known as AdaMax \cite{Adam}, proved to learn the quickest without becoming stuck in local optima. In addition, AdaMax requires relatively little tuning of parameters because it uses estimates of the lower-order moments of the gradient to anneal the step size of the gradient descent \cite{Adam}. 

\begin{figure}[h]
	\centering
	\input{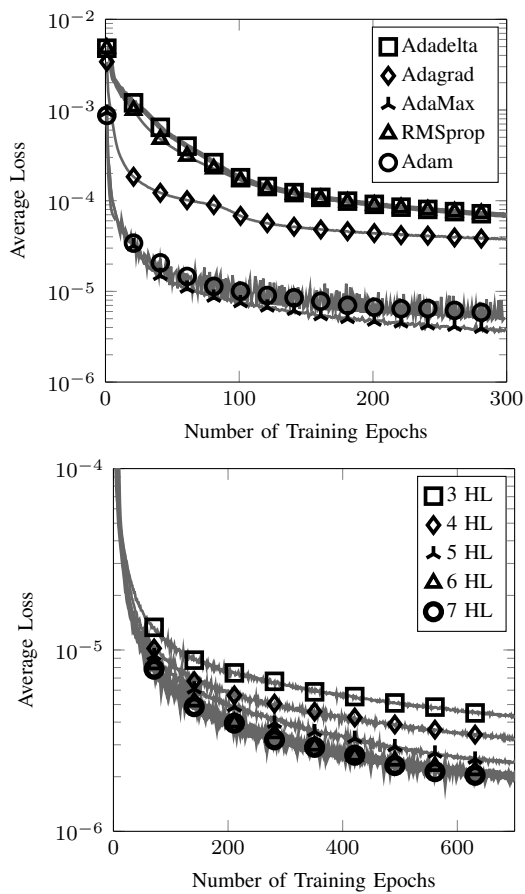}
	\caption{\edit{Training curves for different optimizers (top) and number of hidden layers (bottom)}}
	\label{fig_training}
\end{figure}

After selecting AdaMax for the optimizer, different network architectures were investigated. A baseline architecture was chosen with five hidden layers and a set of layer sizes that is larger early in the network and tapers to smaller layer sizes towards the end of the network. This approach allows for the network to find increasingly more abstract representations of the data, similar to the approach with convolutional neural networks in image classification \cite{krizhevsky2012imagenet}. The layer sizes were also chosen so that the total number of parameters in the network would be around 600,000, as this would mean the table could be compressed to occupy only a few megabytes when using floating point precision. To test the baseline architecture, the number of hidden layers was varied to see the effect on the regression performance, but six hidden layers proved to give the best results with additional layers yielding little improvement. \Cref{fig_training} shows the network loss during training for the different network optimizers and number of layers. As a result of this study, a neural network with six hidden layers and AdaMax optimization was chosen.

\subsection{Implementation}
The deep neural network was trained in Python using the Keras library \cite{keras} running on top of Theano \cite{theano}. To reduce training time, the deep neural network was trained using an NVIDIA DIGITS DevBox with four Titan X GPUs. Before training, the score table and inputs were normalized to have zero mean and unit range, which helps the network train more quickly \cite{LeCun2012}.
For each training epoch, the table data was shuffled and passed through the network in batches of  $2^{16}$ samples. The network was trained for 1200 training epochs over the course of four days.

\subsection{Results}
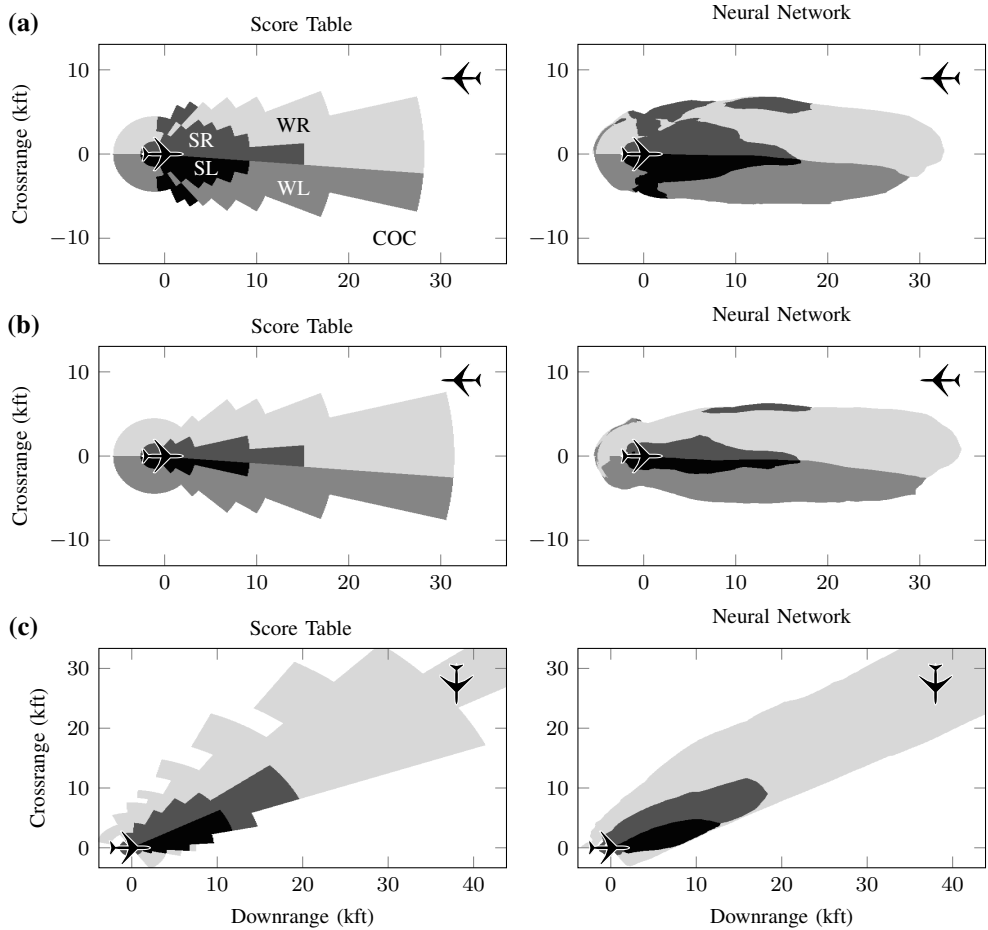
\begin{figure*}[h]
	\centering
	\begin{tikzpicture}[]
\begin{groupplot}[group style={horizontal sep=0.95cm, group size=2 by 3, vertical sep=1.1cm}, width = 7cm, height = 4.5cm]
\nextgroupplot [title = {\textbf{(a)}$\qquad\qquad\qquad\qquad$\footnotesize Score Table},every axis title/.style={above,at={(0.2,1)}},  ylabel = {Crossrange (kft)}, view = {{0}{90}}, enlargelimits = false, axis on top]\addplot [point meta min=-3, point meta max=3] graphics [xmin=-7.142857142857146, xmax=37.142857142857146, ymin=-13.025210084033615, ymax=13.025210084033615] {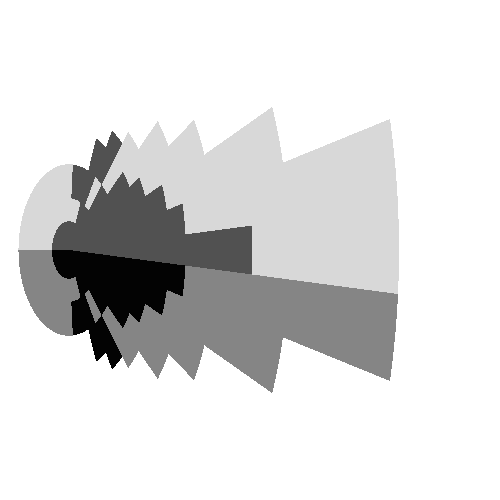};
\node[aircraft top,draw=white,fill=black,minimum width=1cm,rotate=0,scale = 0.55] at (axis cs:0.0, 0.0) {};
\node[aircraft top,draw=white,fill=black,minimum width=1cm,rotate=180,scale = 0.55] at (axis cs:32,9) {};

\node[text=black] at (25,-10)  {\footnotesize COC};
\node[text=white] at (4.5,-1.7)  {\footnotesize SL};
\node[text=white] at (4.,1.7)  {\footnotesize SR};
\node[text=white] at (14,-4)  {\footnotesize WL};
\node[text=black] at (14,3.5)  {\footnotesize WR};

\nextgroupplot [ title = {Neural Network}, view = {{0}{90}}, enlargelimits = false, axis on top]\addplot [point meta min=-3, point meta max=3] graphics [xmin=-7.142857142857146, xmax=37.142857142857146, ymin=-13.025210084033615, ymax=13.025210084033615] {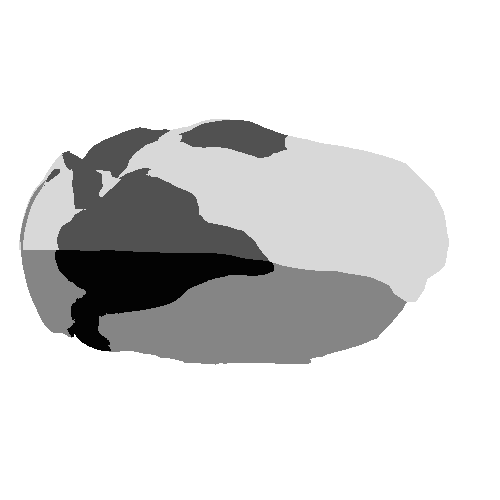};
\node[aircraft top,draw=white,fill=black,minimum width=1cm,rotate=0,scale = 0.55] at (axis cs:0.0, 0.0) {};
\node[aircraft top,draw=white,fill=black,minimum width=1cm,rotate=180,scale = 0.55] at (axis cs:32,9) {};

\nextgroupplot [ title = {\textbf{(b)}$\qquad\qquad\qquad\qquad$\footnotesize Score Table},every axis title/.style={above,at={(0.2,1)}},  ylabel = {Crossrange (kft)}, view = {{0}{90}}, enlargelimits = false, axis on top]\addplot [point meta min=-3, point meta max=3] graphics [xmin=-7.142857142857146, xmax=37.142857142857146, ymin=-13.025210084033615, ymax=13.025210084033615] {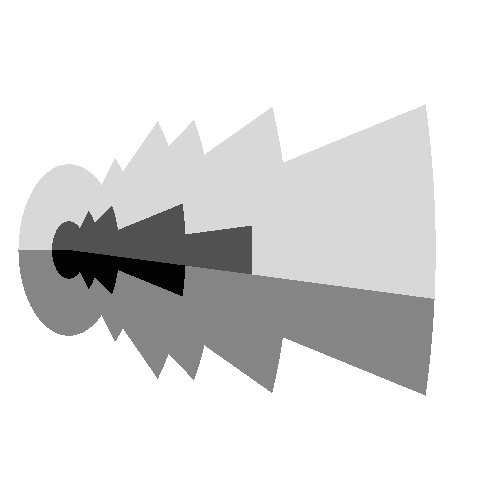};
\node[aircraft top,draw=white,fill=black,minimum width=1cm,rotate=0,scale = 0.55] at (axis cs:0.0, 0.0) {};
\node[aircraft top,draw=white,fill=black,minimum width=1cm,rotate=180,scale = 0.55] at (axis cs:32,9) {};

\nextgroupplot [ title = {Neural Network}, view = {{0}{90}}, enlargelimits = false, axis on top]\addplot [point meta min=-3, point meta max=3] graphics [xmin=-7.142857142857146, xmax=37.142857142857146, ymin=-13.025210084033615, ymax=13.025210084033615] {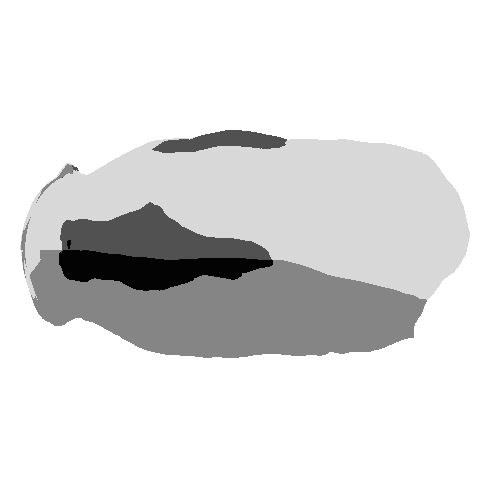};
\node[aircraft top,draw=white,fill=black,minimum width=1cm,rotate=0,scale = 0.55] at (axis cs:0.0, 0.0) {};
\node[aircraft top,draw=white,fill=black,minimum width=1cm,rotate=180,scale = 0.55] at (axis cs:32,9) {};

\nextgroupplot [title = {\textbf{(c)}$\qquad\qquad\qquad\qquad$\footnotesize Score Table},every axis title/.style={above,at={(0.2,1)}}, xlabel = {Downrange (kft)}, ylabel = {Crossrange (kft)}, view = {{0}{90}}, enlargelimits = false, axis on top]\addplot [point meta min=-3, point meta max=3] graphics [xmin=-3.8461538461538467, xmax=43.84615384615385, ymin=-3.3431952662721898, ymax=33.34319526627219] {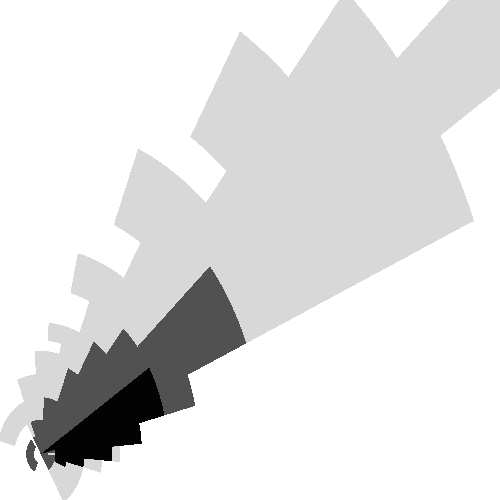};
\node[aircraft top,draw=white,fill=black,minimum width=1cm,rotate=0,scale = 0.55] at (axis cs:0.0, 0.0) {};
\node[aircraft top,draw=white,fill=black,minimum width=1cm,rotate=-90,scale = 0.55] at (axis cs:38,27) {};

\nextgroupplot [title = {Neural Network}, xlabel = {Downrange (kft)}, view = {{0}{90}}, enlargelimits = false, axis on top]\addplot [point meta min=-3, point meta max=3] graphics [xmin=-3.8461538461538467, xmax=43.84615384615385, ymin=-3.3431952662721898, ymax=33.34319526627219] {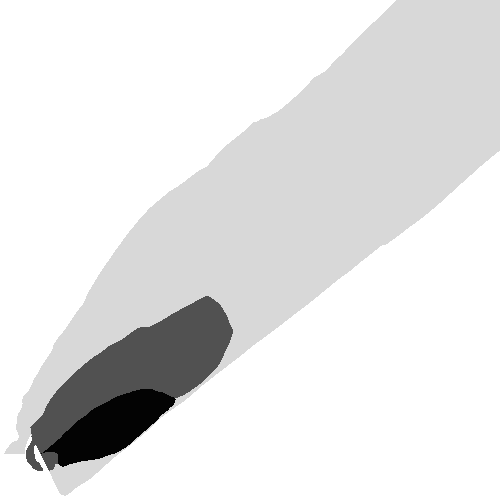};
\node[aircraft top,draw=white,fill=black,minimum width=1cm,rotate=0,scale = 0.55] at (axis cs:0.0, 0.0) {};
\node[aircraft top,draw=white,fill=black,minimum width=1cm,rotate=-90,scale = 0.55] at (axis cs:38,27) {};

\end{groupplot}

\end{tikzpicture}
	\caption{\edit{Policy plots of the original table and neural network representation for (a) head-on encounter with $a_\text{prev}=\text{COC}$ and $\tau=0\text{s}$, (b) head-on encounter with $a_\text{prev}=\text{COC}$ and $\tau=20\text{s}$, and (c) a $90^\circ$ encounter with $a_\text{prev}=\text{WR}$ and $\tau=0\text{s}$}}
	\label{fig_PolicyPlot}
\end{figure*}

\Cref{fig_PolicyPlot} shows the policies for \edit{three} different encounter geometries. \edit{\Cref{fig_PolicyPlot} illustrates that increasing $\tau$ shrinks the strong alerting region because the aircraft have greater vertical separation, and changing $a_\text{prev}$ to WR yields a policy that favors right turns in order to reduce reversals}. Qualitatively the plots show that the neural network is a continuous approximation of the discrete table, which uses nearest-neighbor interpolation. Overall compression quality is computed by evaluating the network at each discrete state in the table and comparing the predicted scores to the table scores. The network's predictions have RMSE of 0.923 with a policy error of 2.02\%, which is lower than all but the largest baseline decision trees shown in \cref{fig_DT_pareto}. Given that the neural network requires only \SI{2.4}{\mega\byte} of floating point storage, the neural network method is an efficient and accurate compression approach.

To evaluate operational performance of the neural network, the network was evaluated in 1.5 million simulated \edit{3D} encounters with varied encounter configurations and sensor noise. Overall performance metrics were computed from the simulation results that quantify the overall safety and efficiency of the system. Four performance metrics were considered with the desire to minimize all four metrics:
\begin{enumerate}
	\item P(NMAC): the probability of a near mid-air collision
	\item P(Alert): the probability that the system will give an alert
	\item P(Reversal): the probability that the system will reverse the direction of its advisory
	\item Relative Runtime: required runtime to compute an advisory, normalized to the original score table
\end{enumerate}

The aggregate results from the simulations are shown in \cref{simcomp}. The neural network outperforms the table in the three performance probability metrics. However, due to the large amount of computation required to compute the score values using the deep neural network, the runtime is increased by a factor of 50. This is a significant increase that is addressed in the next section.

\begin{table} 
	\caption{Simulation comparison}\label{simcomp}
	\centering
	\begin{tabular}{p{1.85cm}p{1.57cm}p{0.93cm}p{1.2cm}p{1.0cm}}
		\toprule
		Representation & P(NMAC) & P(Alert) & P(Reversal)  & Relative Runtime  \\
		\midrule
		Original Table & $1.546\times10^{-4}$ & 0.55485 & 0.007438 & 1$\times$\\
		Neural Network & $1.272\times10^{-4}$ & 0.53128 & 0.006903  & 50$\times$ \\
		\bottomrule
	\end{tabular}
\end{table}

In addition the score table was modified to optimize for a new parameter: P(Split), the probability that the turning alerts given by the system will be split by COC advisories. This behavior is undesirable because the system alerted that the encounter was clear of conflict before the threat was finished. It was found that the existing score table had issues with large numbers of split advisories, so the values were updated to address this issue. As a result, the neural networks presented in the remainder of this work use the updated table and optimize for P(Split).

\section{Improving Network Runtime \label{sec:ImproveRuntime}}
As described in the previous section, a large increase in runtime is a significant drawback of the neural network compression. Although each forward propagation requires \edit{an average of \SI{0.6}{\milli\second} of computation on a modern computer}, this computation will be performed on slower legacy avionics systems. In addition, multiple states in a weighted set must be evaluated, and look-aheads to determine duration of an advisory further increase the number of evaluated points. \edit{Furthermore, the system must be able to handle encounters where up to 30 intruders are present, which means hundreds of forward propagations and any additional computation must be able to run within the allotted one second between advisories. To enable the neural network to run in real time in the most stressing scenarios, the increase in computation time must be eliminated}. This section presents two methods for reducing required runtime of the neural network. 

\subsection{Network Pruning \label{sec:NNetPruning}}
The first approach to speed network evaluations reduces the number of computations performed by the network by making the network sparse. Experiments in making computer vision networks sparse show that 90\% of network connections can be removed without degrading network performance \cite{han2015deep}. If similar results can be achieved on the ACAS Xu networks, then runtime required can be reduced by up to a factor of 10, though the actual speedup factor will be lower due to overhead associated with a sparse representation. Although this speedup is not enough to entirely eliminate the runtime increase, this could still contribute significantly to runtime reduction.

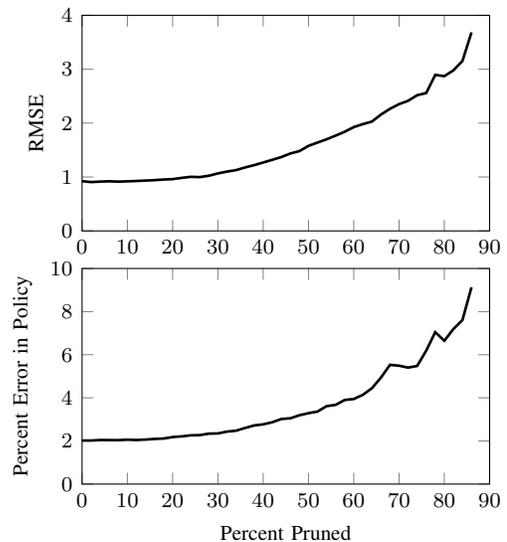
\begin{figure}[h]
	\centering
	\begin{tikzpicture}

\begin{groupplot}[group style={vertical sep = 0.5cm, group size=1 by 2},every x tick label/.append style={font=\footnotesize},every y tick label/.append style={font=\footnotesize}, width=7.0cm,height=4.45cm]
\nextgroupplot[
xmin=0,
xmax=90,
ymin=0.0,
ymax=4.0,
ylabel={RMSE},
xtick={0,10,20,30,40,50,60,70,80,90},
]
\addplot [color=black,solid,forget plot,line width=1.0pt]
  table[row sep=crcr]{%
0	0.9230\\
2	0.9061\\
4	0.9148\\
6	0.9203\\
8	0.9142\\
10	0.9196\\
12	0.9261\\
14	0.9340\\
16	0.9421\\
18	0.9531\\
20	0.9609\\
22	0.9837\\
24	1.0035\\
26	0.9982\\
28	1.0235\\
30	1.0664\\
32	1.1004\\
34	1.1277\\
36	1.1773\\
38	1.2201\\
40	1.2688\\
42	1.3187\\
44	1.3699\\
46	1.4362\\
48	1.4804\\
50	1.5779\\
52	1.6393\\
54	1.7000\\
56	1.7677\\
58	1.8402\\
60	1.9258\\
62	1.9812\\
64	2.0283\\
66	2.1581\\
68	2.2653\\
70	2.3521\\
72	2.4138\\
74	2.5176\\
76	2.5584\\
78	2.8948\\
80	2.8708\\
82	2.9782\\
84	3.1525\\
86	3.6829\\
};

\nextgroupplot[%
xmin=0,
xmax=90,
xlabel={Percent Pruned},
ymin=0.0,
ymax=10.0,
ylabel={Percent Error in Policy},
xtick={0,10,20,30,40,50,60,70,80,90},
]
\addplot [color=black,solid,forget plot,line width=1.0pt]
table[row sep=crcr]{%
	0	2.0200\\
	2	2.0205\\
	4	2.0444\\
	6	2.0417\\
	8	2.0379\\
	10	2.0588\\
	12	2.0442\\
	14	2.0647\\
	16	2.0944\\
	18	2.1126\\
	20	2.1818\\
	22	2.2128\\
	24	2.2613\\
	26	2.2718\\
	28	2.3406\\
	30	2.3486\\
	32	2.4344\\
	34	2.4760\\
	36	2.6029\\
	38	2.7192\\
	40	2.7711\\
	42	2.8681\\
	44	3.0191\\
	46	3.0519\\
	48	3.1933\\
	50	3.2891\\
	52	3.3654\\
	54	3.6159\\
	56	3.6744\\
	58	3.9023\\
	60	3.9465\\
	62	4.1400\\
	64	4.4449\\
	66	4.9392\\
	68	5.5305\\
	70	5.4912\\
	72	5.4031\\
	74	5.4786\\
	76	6.1903\\
	78	7.0476\\
	80	6.6465\\
	82	7.1919\\
	84	7.6071\\
	86	9.1282\\
};

\end{groupplot}
\end{tikzpicture}%
	\caption{\edit{RMSE of network predictions (top) and policy error rate (bottom) for pruned networks}}
	\label{fig_prune}
\end{figure}

\begin{figure*}[h]
\centering
\begin{tikzpicture}[]

\begin{groupplot}[group style={horizontal sep=1.0cm, vertical sep = 1.4cm, group size=3 by 2},every x tick label/.append style={font=\footnotesize},every y tick label/.append style={font=\footnotesize}, width=6.0cm,height=4.3cm]

\nextgroupplot [ ylabel = {Crossrange (kft)}, enlargelimits = false, axis on top, title={\textbf{(a)}},every axis title/.style={above,at={(0,1)}} ]\addplot [point meta min=-3, point meta max=3] graphics [xmin=-11, xmax=51, ymin=-12, ymax=12] {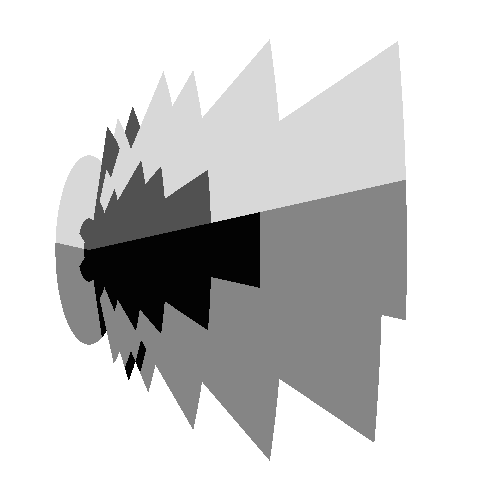};
\node[aircraft top,draw=white,fill=black,minimum width=1cm,rotate=0,scale = 0.55] at (axis cs:0.0, 0.0) {};
\node[aircraft top,fill=black,minimum width=1cm,rotate=-180,scale = 0.55] at (axis cs:44.5,9.25) {};

\node[text=black] at (43,-7)  {\scriptsize COC};
\node[text=white] at (10,-1.7)  {\scriptsize SL};
\node[text=white] at (10,1.7)  {\scriptsize SR};
\node[text=white] at (30,-2)  {\scriptsize WL};
\node[text=black] at (30,4.5)  {\scriptsize WR};

\nextgroupplot [enlargelimits = false, axis on top, title={\textbf{(b)}},every axis title/.style={above,at={(0,1)}}]\addplot [point meta min=-3, point meta max=3] graphics [xmin=-11, xmax=51, ymin=-12, ymax=12] {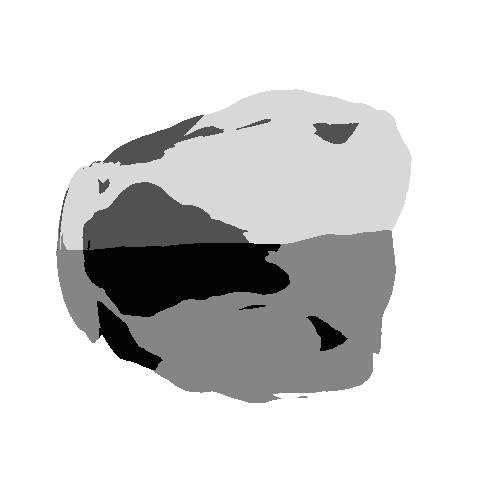};
\node[aircraft top,draw=white,fill=black,minimum width=1cm,rotate=0,scale = 0.55] at (axis cs:0.0, 0.0) {};
\node[aircraft top,fill=black,minimum width=1cm,rotate=-180,scale = 0.55] at (axis cs:44.5,9.25) {};

\nextgroupplot [enlargelimits = false, axis on top, title={\textbf{(c)}},every axis title/.style={above,at={(0,1)}}]\addplot [point meta min=-3, point meta max=3] graphics [xmin=-11, xmax=51, ymin=-12, ymax=12] {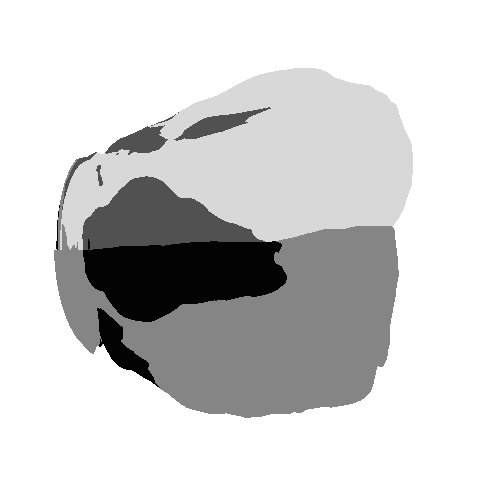};
\node[aircraft top,draw=white,fill=black,minimum width=1cm,rotate=0,scale = 0.55] at (axis cs:0.0, 0.0) {};
\node[aircraft top,fill=black,minimum width=1cm,rotate=-180,scale = 0.55] at (axis cs:44.5,9.25) {};

\nextgroupplot [xlabel = {Downrange (kft)}, ylabel = {Crossrange (kft)}, enlargelimits = false, axis on top, title={\textbf{(d)}},every axis title/.style={above,at={(0,1)}}]\addplot [point meta min=-3, point meta max=3] graphics [xmin=-11, xmax=51, ymin=-12, ymax=12] {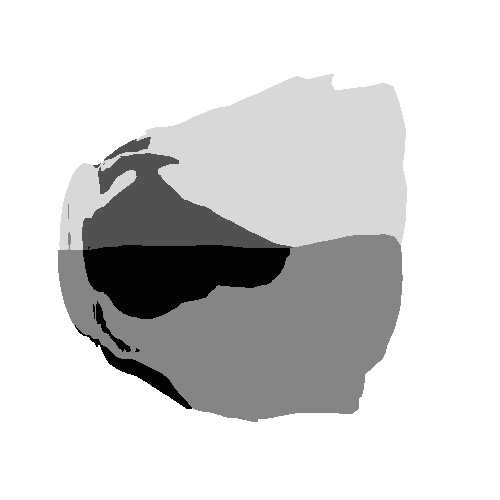};
\node[aircraft top,draw=white,fill=black,minimum width=1cm,rotate=0,scale = 0.55] at (axis cs:0.0, 0.0) {};
\node[aircraft top,fill=black,minimum width=1cm,rotate=-180,scale = 0.55] at (axis cs:44.5,9.25) {};

\nextgroupplot [xlabel = {Downrange (kft)}, enlargelimits = false, axis on top, title={\textbf{(e)}},every axis title/.style={above,at={(0,1)}}]\addplot [point meta min=-3, point meta max=3] graphics [xmin=-11, xmax=51, ymin=-12, ymax=12] {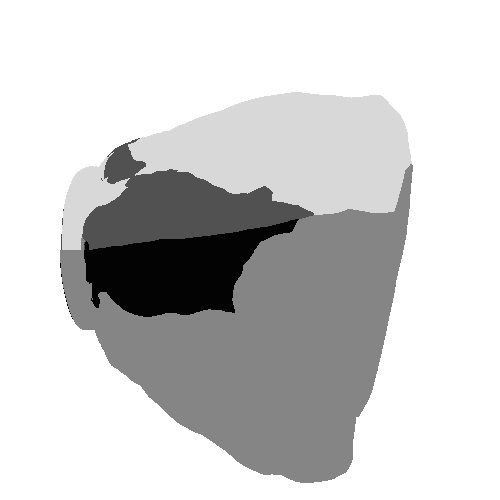};
\node[aircraft top,draw=white,fill=black,minimum width=1cm,rotate=0,scale = 0.55] at (axis cs:0.0, 0.0) {};
\node[aircraft top,fill=black,minimum width=1cm,rotate=-180,scale = 0.55] at (axis cs:44.5,9.25) {};

\nextgroupplot [xlabel = {Downrange (kft)}, enlargelimits = false, axis on top, title={\textbf{(f)}},every axis title/.style={above,at={(0,1)}}]\addplot [point meta min=-3, point meta max=3] graphics [xmin=-11, xmax=51, ymin=-12, ymax=12] {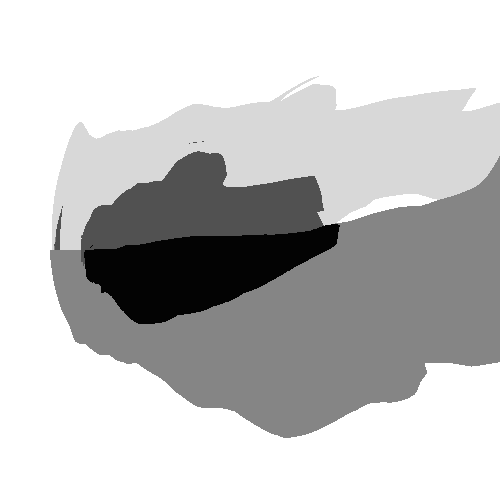};
\node[aircraft top,draw=white,fill=black,minimum width=1cm,rotate=0,scale = 0.55] at (axis cs:0.0, 0.0) {};
\node[aircraft top,fill=black,minimum width=1cm,rotate=-180,scale = 0.55] at (axis cs:44.5,9.25) {};

\end{groupplot}

\end{tikzpicture}
\caption{\edit{Policy plots for (a) the original table as well as the neural networks with (b) 0\%, (c) 20\%, (d) 40\%, (e) 60\%, and (f) 80\% of connections pruned}}
\label{fig_prunePolicy}
\end{figure*}
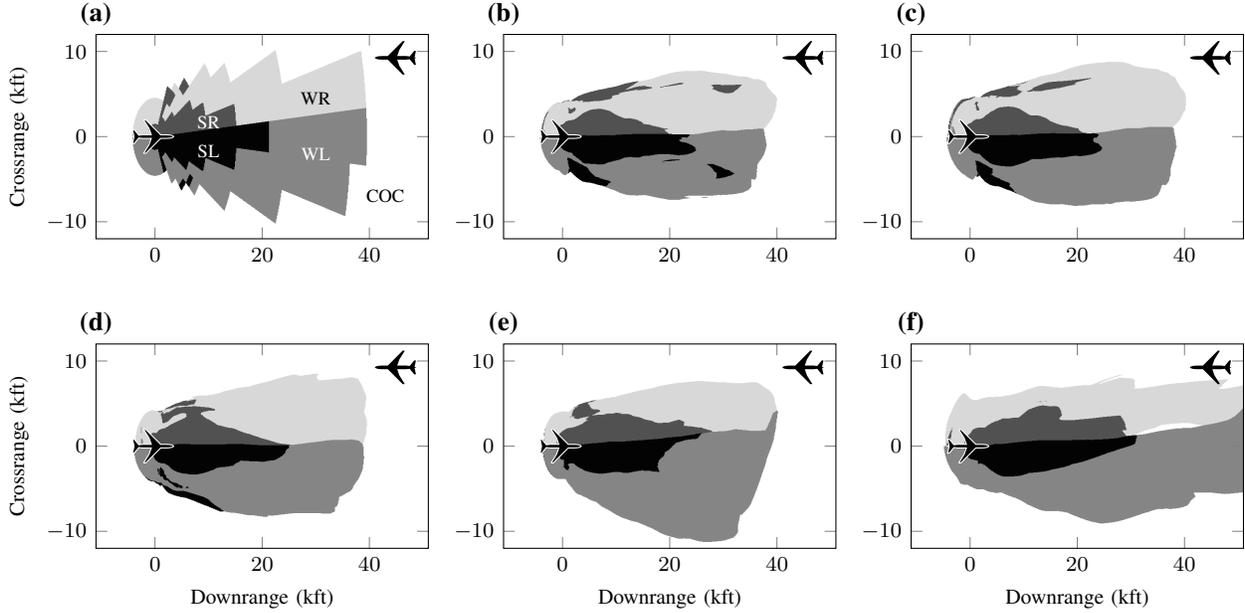

Pruning the networks is an iterative process that removes a small fraction of the network connections and then retrains the network to adjust the remaining connections. Removing too many connections at once could have a serious impact on the network's performance, so this implementation removes only 2\% of the network connections at a time. In addition, the network's least important connections should be removed first. One metric for identifying unimportant connections is the magnitude of the connection's weight. Connections with low magnitude weight likely have little impact on the network's output, so fixing the weight to zero will have the smallest impact on the network output.

\edit{The RMSE and policy error rates are plotted as a function of pruning in \cref{fig_prune}. With 60\% of connections pruned, the RMSE and policy error rate have both doubled, and further pruning increases the network errors.} In addition, plotting the pruned policies demonstrates how pruning affects the network. \Cref{fig_prunePolicy} shows the policy plot of a head-on encounter. \edit{The policy changes seen in \cref{fig_prune} are sporadic and global, which is expected for a neural network where weight changes in early layers can have an impact on all values in future layers.} While pruning 40\% of the connections yields only minor policy changes, there are more significant changes when 60\% is pruned. If pruning increases to 80\%, the policy changes drastically, and likely the performance of the 80\% pruned network will be degraded. Therefore, the network can be pruned up to 60\% of the network weights before seeing major changes in the policy. This result suggests that the ACAS Xu networks are dense compared to computer vision networks. Due to the additional overhead for using a sparse representation, pruning only 60\% of weights will not achieve much speedup, so another approach will be required.

\subsection{Multiple Small Neural Networks \label{sec:NNetArray}}

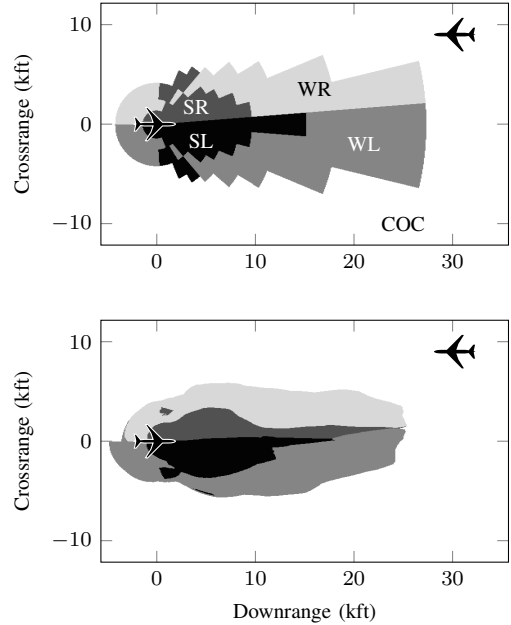
\begin{figure}[h]
	\centering
	\begin{tikzpicture}[]
\begin{groupplot}[group style={horizontal sep=1.2cm, vertical sep=1.0cm, group size=1 by 2}]
\nextgroupplot [height = {4.8cm}, ylabel = {Crossrange (kft)}, width = {7cm}, enlargelimits = false, axis on top]\addplot [point meta min=-3, point meta max=4.5] graphics  [xmin=-5.666666666666668, xmax=35.66666666666667, ymin=-12.15686274509804, ymax=12.15686274509804] {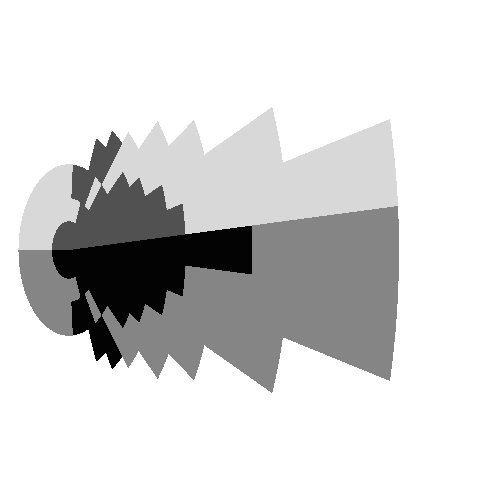};
\node[aircraft top,draw=white,fill=black,minimum width=1cm,rotate=0,scale = 0.55] at (axis cs:0.0, 0.0) {};
\node[aircraft top,fill=black,minimum width=1cm,rotate=-180,scale = 0.55] at (axis cs:30,9) {};
\node[text=black] at (25,-10)  {\footnotesize COC};
\node[text=white] at (4.5,-1.7)  {\footnotesize SL};
\node[text=white] at (4.,1.7)  {\footnotesize SR};
\node[text=white] at (21,-2)  {\footnotesize WL};
\node[text=black] at (16,3.5)  {\footnotesize WR};

\nextgroupplot [ ylabel = {Crossrange (kft)},height = {4.8cm}, xlabel = {Downrange (kft)}, width = {7cm}, enlargelimits = false, axis on top]\addplot [point meta min=-3, point meta max=4.5] graphics  [xmin=-5.666666666666668, xmax=35.66666666666667, ymin=-12.15686274509804, ymax=12.15686274509804] {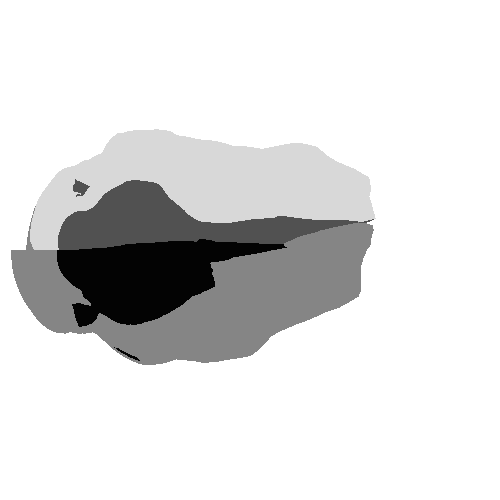};
\node[aircraft top,draw=white,fill=black,minimum width=1cm,rotate=0,scale = 0.55] at (axis cs:0.0, 0.0) {};
\node[aircraft top,fill=black,minimum width=1cm,rotate=-180,scale = 0.55] at (axis cs:30,9) {};
\end{groupplot}

\end{tikzpicture}
	\caption{\edit{Policy plots of original table (top) and smaller neural network (bottom)}}
	\label{fig_SmallNetwork}
\end{figure}

The second method for speeding up network computation uses an array of smaller networks rather than one large network. The training data of the neural network can be divided such that each network approximates distinct parts of the state space. Only one of the small networks is used to evaluate each state, which leads to a runtime speedup. The training data was split into 45 separate datasets by using the 45 different combinations of $\tau$ and $a_\text{prev}$, and a separate network was trained on each dataset. Because the networks are trained on a factor of 45 less data, the networks can be made smaller, and smaller networks require less computation. To fully eliminate the fifty-fold runtime increase, the networks were made approximately 50 times smaller. Some degradation to performance might be expected, since the networks can no longer generalize between $\tau$ and $a_\text{prev}$, but as shown in this paper, performance does not degrade much. An example policy plot is shown in \cref{fig_SmallNetwork}, which shows that the smaller networks can approximate the table well.

\begin{figure*}[h]
	\centering
	\input{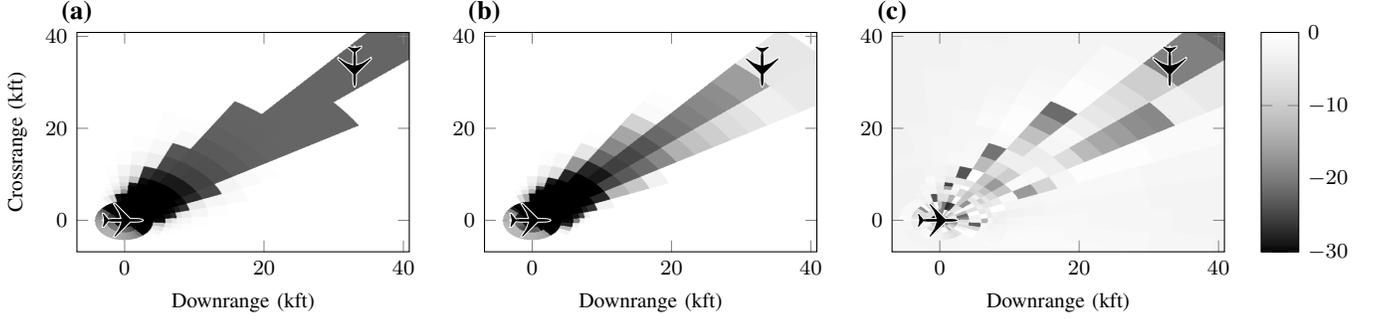}
	\caption{\edit{With the COC penalty included, (a) shows the COC scores for the original table, (b) shows the neural network scores evaluated at the table cutpoints, and (c) shows the COC score errors}}
	\label{fig_trainedWithPenalty}
\end{figure*}

The small networks have 6 hidden layers of 45 hidden units each, giving each network approximately 11000 parameters. Since the total number of parameters in all 45 networks is approximately the same as the single large network, this approach will not increase the memory or storage requirement. Furthermore, training time required for each network is much smaller than training a single large network. Training each network requires approximately 2 hours, and because each network is trained separately, the networks can be trained in parallel on all available GPUs. With four Titan X GPUs, training a new batch of networks can be done in a day, which is also an improvement over training a single large network. This method results in an overall runtime speed up of \edit{3}\% over the original table runtime, so the network pruning technique was not used.

\section{Improving Network Training \label{sec:SmoothData}}
After training the array of small networks, simulations were conducted to compute the overall performance metrics. As reported in \cref{sec:Results}, the performance of the network in simulation shows more work is required to improve the neural network compression. This section presents two methods for improving the quality of the network training data. These adjustments do not change the overall score values the network attempts to represent, but instead finds ways to make the data easier to represent in a neural network format.

\begin{figure*}[h]
	\centering
	\input{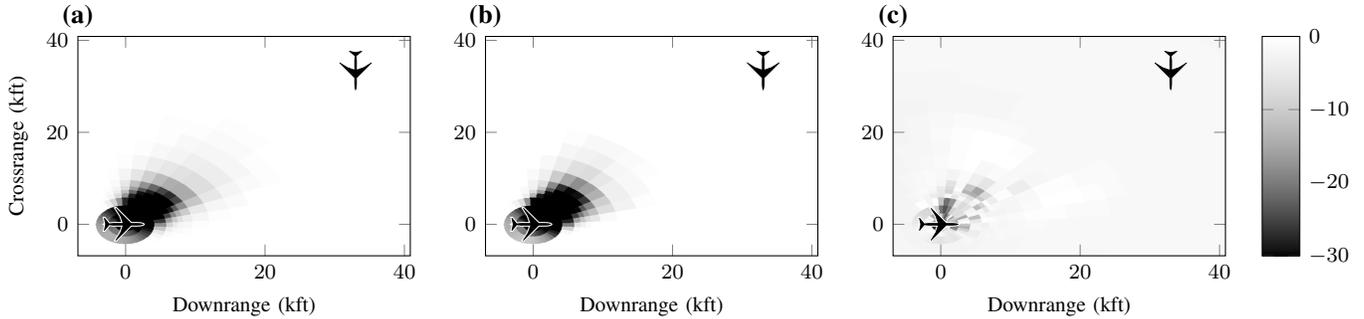}
	\caption{\edit{With the COC penalty removed, (a) shows the COC scores for the original table, (b) shows the neural network scores evaluated at the table cutpoints, and (c) shows the COC score errors}}
	\label{fig_trainedWithoutPenalty}
\end{figure*}

\subsection{Removing Table Discontinuity}
Although neural networks work well as universal function approximators, for a limited network size, some target data is easier to regress. For example, using a neural network to regress data with large discontinuities will generate more loss than smooth target data because networks represent continuous functions. If large discontinuities are present, more connection weight and neuron activations will be required to produce the large jump, and this might disrupt the network's performance for other areas of the state space. Removing such discontinuities will allow the neural network to achieve better performance on the entire data set.

When regressing the score table, large jumps in score values were found that make network regression more error prone. When $a_\text{prev}$ is not COC, and projected straight trajectories of both aircraft results in minimum separation under 4000ft at the closest-point-of-approach (CPA), an additional penalty is added to the COC action. This penalty prevents the system from ending alerts too early when the aircraft are still on course for collision. While other penalty functions used to generate the score table spread to nearby states and create smooth changes in score values, this penalty does not. If the COC action is taken, then the next time step's $a_\text{prev}$ will be COC, and the penalty will not be applied again. As a result, the penalty does not spread to neighboring states and remains within the region in which it is applied.

\Cref{fig_trainedWithPenalty} plots the original table's COC score value in a similar plot to the previous policy plots. The figure is a top-down plot with the ownship at the origin and the intruder traveling perpendicular to the ownship. At each point in the first two plots, the value of the COC action is shown. The original table has a dark band extending outwards, which represents the states where this COC penalty was applied. In states far away from the intruder, the COC score outside the band is approximately zero, while the values in the band are at most -15.0. The result is a large discontinuity in the value function that is problematic for neural network regression. As shown in the middle plot in \cref{fig_trainedWithPenalty}, the neural network does not represent the band well, especially far away from the ownship. The rightmost plot shows the error in the neural network representation, and it is clear there are large errors throughout the plot.

One solution to this issue is to remove the COC penalty from the network training data and apply the penalty to the network output values. Because the penalty does not spread to other states in the score table, the penalty can be easily removed and reapplied by computing the minimum distance at CPA ($d_\text{CPA}$) and time to CPA ($t_\text{CPA}$) as 

\begin{align}
dx &= v_\text{int} \cos(\psi) - v_\text{own} \\
dy &= v_\text{int} \sin(\psi) \\
t_\text{CPA} &= (-r \cos(\theta) dx - r \sin(\theta) dy) / (dx^2 + dy^2) \\
x' &= r \cos(\theta) + t_\text{CPA}dx \\
y' &= r \sin(\theta) + t_\text{CPA}dy \\
d_\text{CPA} &= \sqrt{x'^2 + y'^2} 
\end{align}

The COC score when the penalty is removed is shown in the left plot of \cref{fig_trainedWithoutPenalty}, which shows that the band of low score values seen in \cref{fig_trainedWithPenalty} has been removed. As a result, the network trained to represent this score table is a better fit, as shown in the rightmost plot of \cref{fig_trainedWithoutPenalty}. Even the neural network values at close range are more accurate. Because the training data is more easily represented in some parts of the state space and neural networks are global approximators, the entire score table is regressed more accurately.

\subsection{Training with Online Costs}
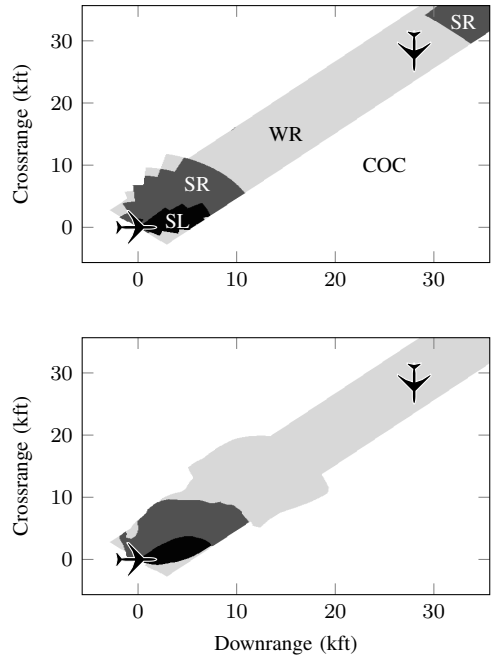
\begin{figure}[h]
	\centering
	\begin{tikzpicture}[]
\begin{groupplot}[group style={vertical sep=1.0cm, group size=1 by 2}, width=7cm, height = 5cm]
\nextgroupplot [ ylabel = {Crossrange (kft)}, enlargelimits = false, axis on top]
\addplot [point meta min=-3, point meta max=3] graphics [xmin=-5.666666666666668, xmax=35.66666666666667, ymin=-5.666666666666668, ymax=35.66666666666667] {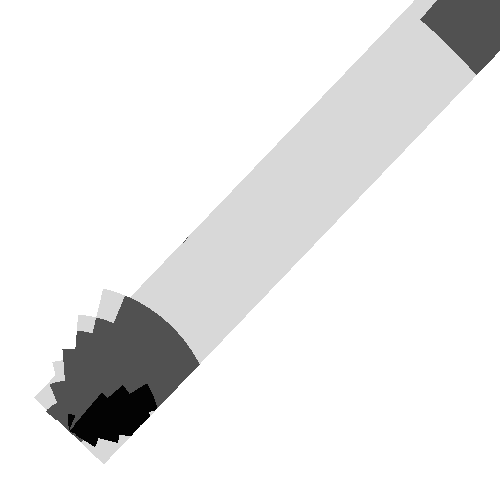};
\node[aircraft top,draw=white,fill=black,minimum width=1cm,rotate=0,scale = 0.55] at (axis cs:0.0, 0.0) {};
\node[aircraft top,draw=white,fill=black,minimum width=1cm,rotate=-90,scale = 0.55] at (axis cs:28,28) {};

\node[text=black] at (25,10)  {\footnotesize COC};
\node[text=white] at (4,1)  {\footnotesize SL};
\node[text=white] at (6,7)  {\footnotesize SR};
\node[text=white] at (33,33)  {\footnotesize SR};
\node[text=black] at (15,15)  {\footnotesize WR};

\nextgroupplot [ylabel = {Crossrange (kft)},xlabel = {Downrange (kft)}, enlargelimits = false, axis on top]
\addplot [point meta min=-3, point meta max=3] graphics [xmin=-5.666666666666668, xmax=35.66666666666667, ymin=-5.666666666666668, ymax=35.66666666666667] {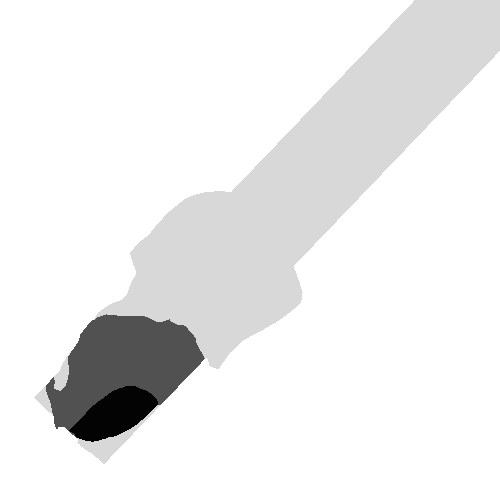};
\node[aircraft top,draw=white,fill=black,minimum width=1cm,rotate=0,scale = 0.55] at (axis cs:0.0, 0.0) {};
\node[aircraft top,draw=white,fill=black,minimum width=1cm,rotate=-90,scale = 0.55] at (axis cs:28,28) {};
\end{groupplot}

\end{tikzpicture}
	\caption{Policy plots with COC penalty added to COC score \edit{for the original table (top) and the neural network (bottom)}}
	\label{fig_PolicySmooth}
\end{figure}

\begin{figure*}[h]
	\centering
	\begin{tikzpicture}[]
\begin{groupplot}[group style={horizontal sep=1.0cm, group size=3 by 1},width=6.5cm, height = 5.1cm]
\nextgroupplot [title={\textbf{(a)}},every axis title/.style={above,at={(0,1)}},  xlabel = {Downrange (kft)}, ylabel = {Crossrange (kft)}, enlargelimits = false, axis on top]
\addplot [point meta min=-3, point meta max=3] graphics [xmin=-5.666666666666668, xmax=35.66666666666667, ymin=-5.666666666666668, ymax=35.66666666666667] {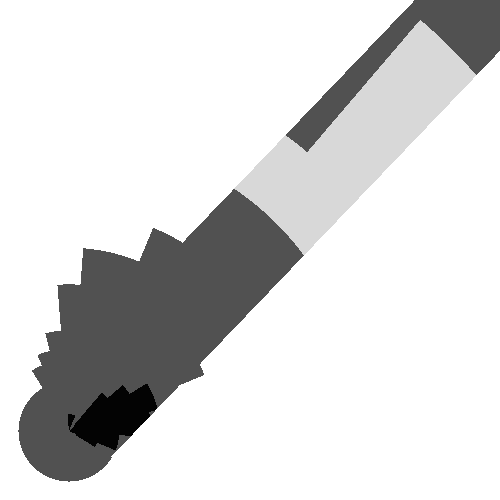};
\node[aircraft top,draw=white,fill=black,minimum width=1cm,rotate=0,scale = 0.55] at (axis cs:0.0, 0.0) {};
\node[aircraft top,draw=white,fill=black,minimum width=1cm,rotate=-90,scale = 0.55] at (axis cs:28,28) {};

\node[text=black] at (25,10)  {\footnotesize COC};
\node[text=white] at (4,1)  {\footnotesize SL};
\node[text=white] at (6,7)  {\footnotesize SR};
\node[text=white] at (33,33)  {\footnotesize SR};
\node[text=black] at (20,20)  {\footnotesize WR};

\nextgroupplot [title={\textbf{(b)}},every axis title/.style={above,at={(0,1)}},  xlabel = {Downrange (kft)}, enlargelimits = false, axis on top]
\addplot [point meta min=-3, point meta max=3] graphics [xmin=-5.666666666666668, xmax=35.66666666666667, ymin=-5.666666666666668, ymax=35.66666666666667] {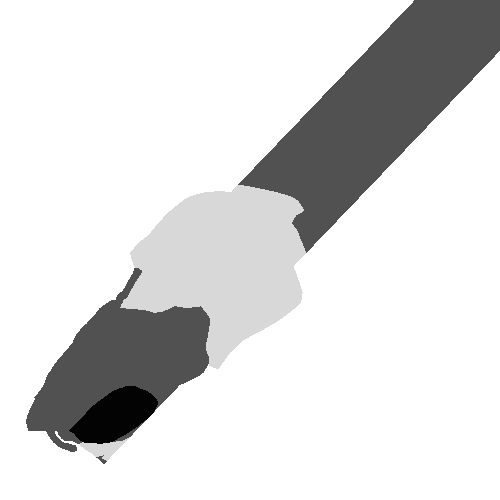};
\node[aircraft top,draw=white,fill=black,minimum width=1cm,rotate=0,scale = 0.55] at (axis cs:0.0, 0.0) {};
\node[aircraft top,draw=white,fill=black,minimum width=1cm,rotate=-90,scale = 0.55] at (axis cs:28,28) {};

\nextgroupplot [title={\textbf{(c)}},every axis title/.style={above,at={(0,1)}},  xlabel = {Downrange (kft)}, enlargelimits = false, axis on top]
\addplot [point meta min=-3, point meta max=3] graphics [xmin=-5.666666666666668, xmax=35.66666666666667, ymin=-5.666666666666668, ymax=35.66666666666667] {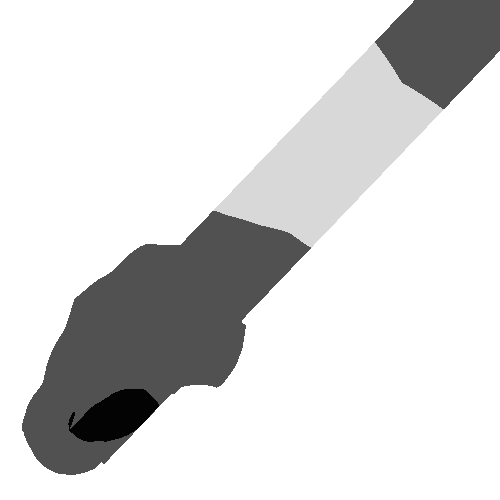};
\node[aircraft top,draw=white,fill=black,minimum width=1cm,rotate=0,scale = 0.55] at (axis cs:0.0, 0.0) {};
\node[aircraft top,draw=white,fill=black,minimum width=1cm,rotate=-90,scale = 0.55] at (axis cs:28,28) {};

\end{groupplot}

\end{tikzpicture}
	\caption{\edit{Policies with online costs added for (a) the original table, (b) the neural network trained without online costs, and (c) the neural network trained with online costs}}
	\label{fig_onlinePolicyPlots}
\end{figure*}
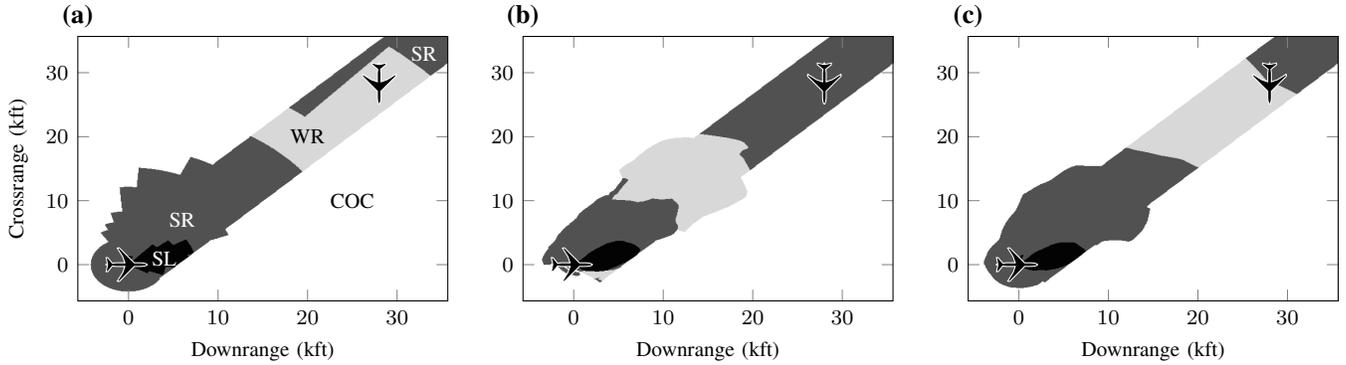

\Cref{fig_PolicySmooth} shows the filtered table and neural network policies when the COC penalty is added back to the COC score. The table and network policies match well, yet the simulation results still show more reversals than expected when using the network. Further investigation revealed that the table values are modified using online costs before being used in the system. The online costs \edit{deterministically} modify the table values used in the collision avoidance system, allowing \edit{a designer to tune the system and optimize} performance. One important online cost decreases the score of COC and weak turning advisories when the previous advisory is a strong turning advisory, which is applied to prevent the system from ending strong alerts too early. If the system stops alerting too early, then the system may need to alert again in the future, which will lead to split advisories and possibly reversals. The left two plots of \cref{fig_onlinePolicyPlots} shows the table and neural network policies when the online costs are applied. While the table's policy changes to increase the area of strong alerts around the ownship, the neural network does not change as much. As a result, \edit{the online costs cannot tune the neural network performance like the original table}, and the neural network is likely to see more reversals and split advisories.

The inability of the neural network to \edit{change its policy with} online costs results from the loss function used to optimize network weights. The asymmetric mean squared error loss encourages the network to under-estimate the scores of suboptimal actions and over-estimate the score of the optimal action. As a result, the score gap between the optimal action and next best action widens, which makes the network policy more difficult to change with online costs. A simple solution to this issue is to calculate the desired policy from the table with online costs added, and then use this policy to determine the optimal action when calculating the network loss. When the network is trained with this approach, the resulting network policy with online costs applied better matches the table policy, as seen in the right plot of \cref{fig_onlinePolicyPlots}.

\section{Results \label{sec:Results}}
\begin{table*} 
	\caption{Simulation Comparison}\label{simcomp2}
	\centering
	\begin{tabular}{lccccc}
		\toprule
		Compression & P(NMAC) & P(Alert) & P(Reversal) & P(Split) & Relative Runtime \\
		\midrule
		Updated Table & $1.472\times10^{-4}$ & 0.6705 & 0.0270 & 0.0969 & 1$\times$ \\
		Nominal Network Array  & $1.846\times10^{-4}$ & 0.5668 & 0.0475 & 0.0701 & 0.97$\times$ \\
		Networks without COC Penalty &  $1.564\times10^{-4}$ & 0.6098 & 0.0435 & 0.0754 & 0.97$\times$  \\
		Networks with Online Costs & $1.451\times10^{-4}$ & 0.6513 & 0.0314 & 0.0801 & 0.97$\times$ \\
		\bottomrule
	\end{tabular}
\end{table*}

To evaluate the neural network with new performance metric measuring the probability of split advisories, P(Split), the table and neural network were simulated with an encounter set of 10 million \edit{3D} encounters with different encounter geometries and sensor noise. \Cref{simcomp2} shows a comparison of the performance metrics for the different neural networks considered in this paper. With each update to the neural network compression method, the overall neural network performance becomes more similar to the updated table. The final version of the neural network compression outperforms the table in all metrics except P(Reversal), for which there is a slight increase. However, the system performs well overall and results in a slightly faster runtime than the original table.

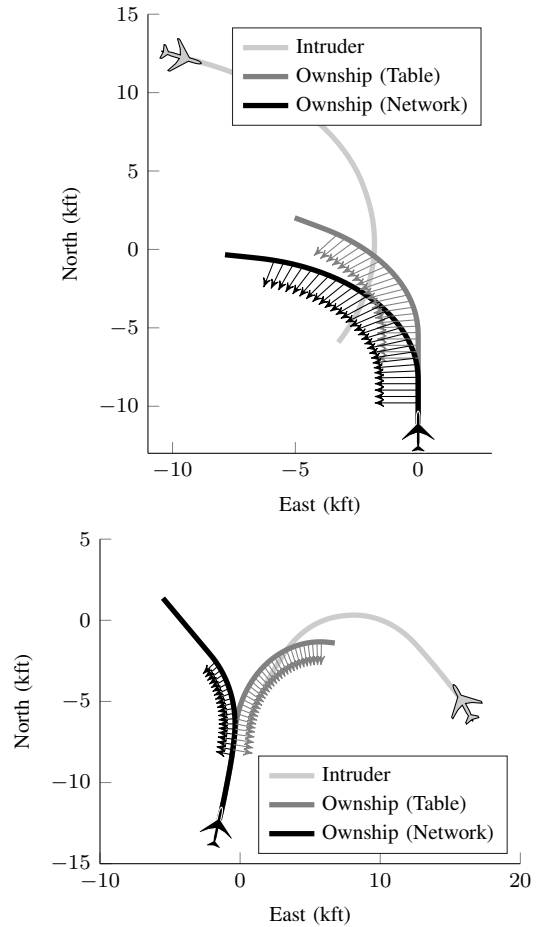
\begin{figure}
	\centering
%
%
\definecolor{mycolor1}{rgb}{0.50000,0.500,0.500}%
\definecolor{mycolor2}{rgb}{0.8000,0.800,0.800}%
\definecolor{mycolor4}{rgb}{0.00000,0.00,0.00}%
\begin{tikzpicture}[%
arrow1/.style={->,color=mycolor1,solid},
arrow2/.style={->,color=mycolor4,solid}
]

\begin{axis}[%
width=1.8in,
height=2.3in,
at={(0in,0in)},
scale only axis,
xmin=-11,
xmax=3,
xlabel={East (kft)},
ymin=-13,
ymax=15,
ylabel={North (kft)},
axis x line*=bottom,
axis y line*=left,
legend style={legend cell align=left,align=left,draw=white!15!black,at={(0.97,0.97)},anchor=north east}
]

\addplot [color=mycolor2,solid, line width=2 ]
table[row sep=crcr]{%
-9.5554 12.2182 \\
-8.9774 11.9956 \\
-8.4122 11.742 \\
-7.8616 11.4583 \\
-7.3271 11.1453 \\
-6.8103 10.8039 \\
-6.3127 10.435 \\
-5.8358 10.0398 \\
-5.3809 9.6193 \\
-4.9494 9.1749 \\
-4.5425 8.7079 \\
-4.1615 8.2195 \\
-3.8074 7.7113 \\
-3.4813 7.1847 \\
-3.1841 6.6412 \\
-2.9167 6.0825 \\
-2.6799 5.5101 \\
-2.4746 4.9257 \\
-2.3016 4.331 \\
-2.1572 3.7286 \\
-2.0315 3.122 \\
-1.929 2.5111 \\
-1.8524 1.8964 \\
-1.8021 1.279 \\
-1.7784 0.65999 \\
-1.7812 0.040556 \\
-1.8106 -0.57819 \\
-1.8665 -1.1951 \\
-1.9489 -1.8091 \\
-2.0575 -2.4189 \\
-2.1921 -3.0235 \\
-2.3526 -3.6218 \\
-2.5386 -4.2127 \\
-2.7498 -4.795 \\
-2.9858 -5.3678 \\
-3.2498 -5.928 \\
};
\addlegendentry{Intruder};

\addplot [color=mycolor1,solid, line width=2 ]
  table[row sep=crcr]{%
0 -11.5 \\
0 -10.1939 \\
0 -9.7862 \\
0 -9.3784 \\
0 -8.9707 \\
0 -8.5629 \\
0 -8.1551 \\
0 -7.7474 \\
0 -7.3396 \\
0 -6.9319 \\
0 -6.5241 \\
0 -6.1164 \\
0 -5.7086 \\
-0.0020681 -5.3009 \\
-0.01817 -4.8935 \\
-0.05471 -4.4874 \\
-0.11238 -4.0838 \\
-0.19109 -3.6837 \\
-0.29063 -3.2884 \\
-0.41073 -2.8987 \\
-0.55105 -2.5159 \\
-0.71122 -2.141 \\
-0.89079 -1.775 \\
-1.0893 -1.4188 \\
-1.3061 -1.0736 \\
-1.5407 -0.74012 \\
-1.7925 -0.41941 \\
-2.0607 -0.11232 \\
-2.3446 0.18031 \\
-2.6434 0.45769 \\
-2.9563 0.71904 \\
-3.2825 0.96366 \\
-3.621 1.1909 \\
-3.9697 1.4023 \\
-4.3217 1.608 \\
-4.6743 1.8127 \\
-5.0271 2.0173 \\
};
\addlegendentry{Ownship (Table)};

\addplot [color=mycolor4,solid, line width=2 ]
  table[row sep=crcr]{%
0 -11.5 \\
0 -10.1939 \\
0 -9.7862 \\
0 -9.3784 \\
0 -8.9707 \\
0 -8.5629 \\
-0.0020681 -8.1552 \\
-0.01817 -7.7478 \\
-0.05471 -7.3417 \\
-0.11238 -6.9381 \\
-0.19109 -6.538 \\
-0.29063 -6.1427 \\
-0.41073 -5.753 \\
-0.55105 -5.3702 \\
-0.71122 -4.9953 \\
-0.89079 -4.6293 \\
-1.0893 -4.2731 \\
-1.3061 -3.9279 \\
-1.5407 -3.5944 \\
-1.7925 -3.2737 \\
-2.0607 -2.9666 \\
-2.3446 -2.674 \\
-2.6434 -2.3966 \\
-2.9563 -2.1353 \\
-3.2825 -1.8906 \\
-3.621 -1.6634 \\
-3.971 -1.4542 \\
-4.3314 -1.2637 \\
-4.7013 -1.0922 \\
-5.0797 -0.94033 \\
-5.4655 -0.80848 \\
-5.8576 -0.69699 \\
-6.2551 -0.60619 \\
-6.6564 -0.53378 \\
-7.0588 -0.46787 \\
-7.4613 -0.40311 \\
-7.864 -0.33856 \\
};
\addlegendentry{Ownship (Network)};

\addplot [arrow1] coordinates{(0,-6.93187247370267) (-1.67561496681106,-6.93187247370267)};
\addplot [arrow1] coordinates{(0,-6.52411526936723) (-1.67561496681106,-6.52411526936723)};
\addplot [arrow1] coordinates{(0,-6.11635806503179) (-1.67561496681106,-6.11635806503179)};
\addplot [arrow1] coordinates{(0,-5.70860086069635) (-1.67561496681106,-5.70860086069635)};
\addplot [arrow1] coordinates{(-0.00206808910336592,-5.30085426296781) (-1.67737681383422,-5.33288852340715)};
\addplot [arrow1] coordinates{(-0.0181704645446716,-4.89345260290037) (-1.69008771623611,-5.00471005432609)};
\addplot [arrow1] coordinates{(-0.0547099554652474,-4.48738141135743) (-1.7185958500744,-4.68529363109614)};
\addplot [arrow1] coordinates{(-0.112382732208322,-4.08376992364426) (-1.76363947699668,-4.36843934559217)};
\addplot [arrow1] coordinates{(-0.19109466994479,-3.68372896359373) (-1.82519087446044,-4.05442440649632)};
\addplot [arrow1] coordinates{(-0.290634872412771,-3.2883556164075) (-1.90309098790768,-3.74406472604244)};
\addplot [arrow1] coordinates{(-0.410730871578088,-2.89873363767961) (-1.99712720748029,-3.43820762448275)};
\addplot [arrow1] coordinates{(-0.551053520010831,-2.51593096051907) (-2.1070418685565,-3.13769118455348)};
\addplot [arrow1] coordinates{(-0.71121820513111,-2.14099682104476) (-2.23253370990024,-2.84333908230366)};
\addplot [arrow1] coordinates{(-0.890785926822452,-1.77495888772432) (-2.3732587677169,-2.55595811500406)};
\addplot [arrow1] coordinates{(-1.08926450245264,-1.41882044509461) (-2.52883132445709,-2.27633597380836)};
\addplot [arrow1] coordinates{(-1.30610991604279,-1.07355764387237) (-2.69882496656086,-2.00523908356174)};
\addplot [arrow1] coordinates{(-1.54072780938705,-0.740116825396879) (-2.8827737532717,-1.7434105019961)};
\addplot [arrow1] coordinates{(-1.79247511114787,-0.419411927774382) (-3.08017349375136,-1.491567883054)};
\addplot [arrow1] coordinates{(-2.06066179946985,-0.112321980837341) (-3.2904831290323,-1.25040150984858)};
\addplot [arrow1] coordinates{(-2.34455279328173,0.180311303215254) (-3.51312621502265,-1.02057240264553)};
\addplot [arrow1] coordinates{(-2.6433699671025,0.457685836892503) (-3.74749250249945,-0.802710507052464)};
\addplot [arrow1] coordinates{(-2.95629428382925,0.719041355877667) (-3.99293960975931,-0.597412967381813)};
\addplot [arrow1] coordinates{(-3.28246803966093,0.963661502859807) (-4.24879478334206,-0.405242489919754)};

\addplot [arrow2] coordinates{(0,-9.78617290405071) (-1.77678791253842,-9.78617290405071)};
\addplot [arrow2] coordinates{(0,-9.37841569971527) (-1.77678791253842,-9.37841569971527)};
\addplot [arrow2] coordinates{(0,-8.97065849537984) (-1.77678791253842,-8.97065849537984)};
\addplot [arrow2] coordinates{(0,-8.56290129104441) (-1.77678791253842,-8.56290129104441)};
\addplot [arrow2] coordinates{(-.00206808910336592,-8.15515469331588) (-1.77853126879015,-8.18912316928626)};
\addplot [arrow2] coordinates{(-.0181704645446716,-7.74775303324844) (-1.79103739543834,-7.86572816404431)};
\addplot [arrow2] coordinates{(-.0547099554652474,-7.3416818417055) (-1.8190605992239,-7.55154392025008)};
\addplot [arrow2] coordinates{(-.112382732208322,-6.93807035399233) (-1.86334168325017,-7.23992799909857)};
\addplot [arrow2] coordinates{(-.19109466994479,-6.5380293939418) (-1.9238569343175,-6.93110727593035)};
\addplot [arrow2] coordinates{(-.290634872412771,-6.14265604675557) (-2.00045042804876,-6.62588068587059)};
\addplot [arrow2] coordinates{(-.410730871578088,-5.75303406802768) (-2.09291316879799,-6.32508127330238)};
\addplot [arrow2] coordinates{(-.551053520010831,-5.37023139086714) (-2.20099180804614,-6.02953324269054)};
\addplot [arrow2] coordinates{(-.71121820513111,-4.99529725139283) (-2.32439011727713,-5.74004665085939)};
\addplot [arrow2] coordinates{(-.890785926822452,-4.62925931807239) (-2.46276987095057,-5.45741495896729)};
\addplot [arrow2] coordinates{(-1.08926450245264,-4.27312087544268) (-2.61575177983519,-5.18241284072768)};
\addplot [arrow2] coordinates{(-1.30610991604279,-3.92785807422044) (-2.78291653115694,-4.91579405782349)};
\addplot [arrow2] coordinates{(-1.54072780938705,-3.59441725574495) (-2.96380593795233,-4.65828939380562)};
\addplot [arrow2] coordinates{(-1.79247511114787,-3.27371235812246) (-3.15792419493083,-4.41060465105918)};
\addplot [arrow2] coordinates{(-2.06066179946985,-2.96662241118541) (-3.36473923744304,-4.17341871624766)};
\addplot [arrow2] coordinates{(-2.34455279328173,-2.6798912713282) (-3.583684199833,-3.94738169953181)};
\addplot [arrow2] coordinates{(-2.6433699671025,-2.39661459345557) (-3.81415896917745,-3.73311315266279)};
\addplot [arrow2] coordinates{(-2.95629428382925,-2.13525907447041) (-4.05553183015374,-3.53120037083396)};
\addplot [arrow2] coordinates{(-3.28246803966093,-1.89063892748827) (-4.30714119652756,-3.3421967829455)};
\addplot [arrow2] coordinates{(-3.62099721500486,-1.66342463932869) (-4.56829742451482,-3.16662043469436)};
\addplot [arrow2] coordinates{(-3.97095392492231,-1.45423898856205) (-4.83828470304736,-3.00495256864692)};
\addplot [arrow2] coordinates{(-4.33137896239658,-1.26365533851678) (-5.1163630157613,-2.85763630518662)};
\addplot [arrow2] coordinates{(-4.70128442745279,-1.09219606573035) (-5.4017701693305,-2.72507542795178)};
\addplot [arrow2] coordinates{(-5.07965643492305,-.94033112815147) (-5.69372388258542,-2.6076332770927)};
\addplot [arrow2] coordinates{(-5.46545789343518,-.808476777018051) (-5.99142393069148,-2.5056317533816)};
\addplot [arrow2] coordinates{(-5.85763134800815,-.696994415941462) (-6.29405433850971,-2.41935043590491)};


\node[aircraft top,draw=white,fill=black,minimum width=1cm,rotate=90,scale = 0.55] at (axis cs:0.0, -11.5000) {};
\node[aircraft top,draw=black,fill=mycolor2,minimum width=1cm,rotate=-17,scale = 0.55] at (axis cs:-9.55544046308333,	12.2181690132706) {};
\end{axis}
\end{tikzpicture}%
%
%
\definecolor{mycolor1}{rgb}{0.50000,0.500,0.500}%
\definecolor{mycolor2}{rgb}{0.8000,0.800,0.800}%
\definecolor{mycolor4}{rgb}{0.00000,0.00,0.00}%
\begin{tikzpicture}[%
arrow1/.style={->,color=mycolor1,solid},
arrow2/.style={->,color=mycolor4,solid}
]

\begin{axis}[%
width=2.2in,
height=1.7in,
ylabel={North (kft)},
at={(0in,0in)},
scale only axis,
xmin=-10,
xmax=20,
xlabel={East (kft)},
ymin=-15,
ymax=5,
axis x line*=bottom,
axis y line*=left,
legend style={legend cell align=left,align=left,draw=white!15!black,at={(0.97,0.03)},anchor=south east}
]

\addplot [color=mycolor2,solid, line width=2]
table[row sep=crcr]{%
	15.8494 -4.9774 \\
	15.5887 -4.674 \\
	15.3265 -4.3741 \\
	15.0629 -4.0777 \\
	14.7978 -3.7849 \\
	14.5314 -3.4956 \\
	14.2636 -3.2099 \\
	13.9945 -2.9277 \\
	13.7242 -2.6491 \\
	13.4527 -2.3741 \\
	13.18 -2.1026 \\
	12.9061 -1.8347 \\
	12.6312 -1.5703 \\
	12.3533 -1.3117 \\
	12.0648 -1.0674 \\
	11.7642 -0.84118 \\
	11.4522 -0.63381 \\
	11.1302 -0.44589 \\
	10.7992 -0.27787 \\
	10.4606 -0.13015 \\
	10.1154 -0.0030571 \\
	9.7651 0.10316 \\
	9.4108 0.18833 \\
	9.0538 0.25235 \\
	8.6954 0.29519 \\
	8.3367 0.31692 \\
	7.9791 0.31883 \\
	7.6235 0.30387 \\
	7.2707 0.27274 \\
	6.9213 0.22569 \\
	6.5763 0.16173 \\
	6.2374 0.07818 \\
	5.9056 -0.024978 \\
	5.5822 -0.1472 \\
	5.2683 -0.28787 \\
	4.9648 -0.44629 \\
	4.6728 -0.62171 \\
	4.3932 -0.81334 \\
	4.127 -1.0203 \\
	3.8748 -1.2417 \\
	3.6376 -1.4766 \\
	3.416 -1.7241 \\
	3.2082 -1.9811 \\
	3.0074 -2.2414 \\
	2.8117 -2.5035 \\
	2.6211 -2.7673 \\
	2.4356 -3.0326 \\
	2.2551 -3.2993 \\
	2.0796 -3.5672 \\
	1.9092 -3.8365 \\
	1.7438 -4.1068 \\
};
\addlegendentry{Intruder};

\addplot [color=mycolor1,solid, line width=2]
  table[row sep=crcr]{%
-1.5953 -12.6416 \\
-1.5023 -12.3358 \\
-1.4132 -12.0288 \\
-1.3287 -11.7205 \\
-1.2485 -11.411 \\
-1.1719 -11.1007 \\
-1.0984 -10.7896 \\
-1.0274 -10.4779 \\
-0.957 -10.1661 \\
-0.88701 -9.8542 \\
-0.81855 -9.5419 \\
-0.75186 -9.2293 \\
-0.68751 -8.9162 \\
-0.62697 -8.6023 \\
-0.57057 -8.2877 \\
-0.51835 -7.9723 \\
-0.47033 -7.6563 \\
-0.42652 -7.3396 \\
-0.38693 -7.0224 \\
-0.34952 -6.7049 \\
-0.30248 -6.3888 \\
-0.23958 -6.0754 \\
-0.1604 -5.7657 \\
-0.065127 -5.4606 \\
0.045984 -5.1609 \\
0.17263 -4.8675 \\
0.31446 -4.581 \\
0.47109 -4.3024 \\
0.64208 -4.0324 \\
0.82697 -3.7716 \\
1.0253 -3.5209 \\
1.2364 -3.281 \\
1.4598 -3.0524 \\
1.6949 -2.8358 \\
1.9409 -2.6318 \\
2.1973 -2.441 \\
2.4634 -2.2638 \\
2.7383 -2.1008 \\
3.0214 -1.9525 \\
3.3119 -1.8191 \\
3.609 -1.7012 \\
3.9118 -1.5989 \\
4.2196 -1.5126 \\
4.5315 -1.4426 \\
4.8466 -1.389 \\
5.164 -1.3519 \\
5.483 -1.3315 \\
5.8026 -1.3279 \\
6.122 -1.3409 \\
6.4405 -1.3684 \\
6.7586 -1.4 \\
};
\addlegendentry{Ownship (Table)};

\addplot [color=mycolor4,solid, line width=2]
  table[row sep=crcr]{%
-1.5953 -12.6416 \\
-1.5023 -12.3358 \\
-1.4132 -12.0288 \\
-1.3287 -11.7205 \\
-1.2485 -11.411 \\
-1.1719 -11.1007 \\
-1.0984 -10.7896 \\
-1.0274 -10.4779 \\
-0.957 -10.1661 \\
-0.88701 -9.8542 \\
-0.81855 -9.5419 \\
-0.75186 -9.2293 \\
-0.68751 -8.9162 \\
-0.62697 -8.6023 \\
-0.57057 -8.2877 \\
-0.51835 -7.9723 \\
-0.47033 -7.6563 \\
-0.42652 -7.3396 \\
-0.38887 -7.0222 \\
-0.36543 -6.7034 \\
-0.35857 -6.3838 \\
-0.36842 -6.0644 \\
-0.39499 -5.7458 \\
-0.43818 -5.4291 \\
-0.4979 -5.1151 \\
-0.57396 -4.8047 \\
-0.66617 -4.4987 \\
-0.77427 -4.1979 \\
-0.89796 -3.9031 \\
-1.0369 -3.6153 \\
-1.1907 -3.3351 \\
-1.359 -3.0634 \\
-1.5413 -2.8008 \\
-1.737 -2.5481 \\
-1.9457 -2.306 \\
-2.1652 -2.0736 \\
-2.3876 -1.844 \\
-2.6104 -1.6148 \\
-2.8332 -1.3855 \\
-3.0559 -1.1563 \\
-3.2787 -0.92702 \\
-3.5015 -0.69777 \\
-3.7243 -0.46853 \\
-3.9471 -0.23928 \\
-4.1699 -0.010041 \\
-4.3927 0.2192 \\
-4.6154 0.44845 \\
-4.8382 0.67769 \\
-5.061 0.90693 \\
-5.2838 1.1362 \\
-5.5066 1.3654 \\
};
\addlegendentry{Ownship (Network)};

\addplot [arrow1] coordinates{(-.51834960185223,-7.97228699362178) (0.90081592309735,-8.19660772070255)};
\addplot [arrow1] coordinates{(-.47033195608857,-7.65625026924561) (0.95169123183268,-7.86167818560007)};
\addplot [arrow1] coordinates{(-.42652146510931,-7.33960291605671) (0.99810793092566,-7.52610133362774)};
\addplot [arrow1] coordinates{(-.38692591222467,-7.02240093373738) (1.04005772431806,-7.18993683091397)};
\addplot [arrow1] coordinates{(-.34952363389963,-6.70493197927231) (1.07557957182667,-6.88777461333804)};
\addplot [arrow1] coordinates{(-.30247809635004,-6.38877620026872) (1.11243504353493,-6.63851883256107)};
\addplot [arrow1] coordinates{(-.23958066776334,-6.07539550117489) (1.16041804236142,-6.3984344886864)};
\addplot [arrow1] coordinates{(-.16040229362621,-5.76572780016843) (1.22077730474053,-6.16157308921322)};
\addplot [arrow1] coordinates{(-.06512742107319,-5.46062784018862) (1.29344275150757,-5.92821490371515)};
\addplot [arrow1] coordinates{(.045184489427414,-5.16093227868532) (1.37822119513796,-5.69898054395618)};
\addplot [arrow1] coordinates{(.172628974155181,-4.86746258571425) (1.47488062661111,-5.4744973536471)};
\addplot [arrow1] coordinates{(.314458913796566,-4.5810231428577)  (1.58315613064569,-5.25538057609528)};
\addplot [arrow1] coordinates{(.471085562570774,-4.30239906081109) (1.70275093286358,-5.04223079264271)};
\addplot [arrow1] coordinates{(.642079617731943,-4.03235402882182) (1.83333723229676,-4.83563223190269)};
\addplot [arrow1] coordinates{(.826972396461121,-3.77162822155228) (1.97455710097951,-4.6361511661633)};
\addplot [arrow1] coordinates{(1.02525712050358,-3.52093627031738) (2.12602346506809,-4.44433435915966)};
\addplot [arrow1] coordinates{(1.23639030521331,-3.28096530433009) (2.28732116578683,-4.26070756742672)};
\addplot [arrow1] coordinates{(1.45979324920673,-3.05237306732799) (2.45800809734972,-4.08577409923687)};
\addplot [arrow1] coordinates{(1.69485362054289,-2.83578611474345) (2.63761641874167,-3.92001343506696)};
\addplot [arrow1] coordinates{(1.94092713508281,-2.63179809635874) (2.82565383603816,-3.76387991337558)};
\addplot [arrow1] coordinates{(2.19733932242749,-2.4409681291532)  (3.02160495174869,-3.61780148529298)};
\addplot [arrow1] coordinates{(2.46338737459442,-2.26381926480253) (3.22493267748552,-3.48217854163675)};
\addplot [arrow1] coordinates{(2.73834207236538,-2.10083705603044) (3.43507970608549,-3.35738281546842)};
\addplot [arrow1] coordinates{(3.02144978402565,-1.95246822574241) (3.65147003915042,-3.24375636319897)};
\addplot [arrow1] coordinates{(3.31193453101614,-1.81911944258929) (3.87351056581841,-3.14161062703599)};
\addplot [arrow1] coordinates{(3.60900011483674,-1.70115620631672) (4.10059268843977,-3.05122558134224)};
\addplot [arrow1] coordinates{(3.91183229937086,-1.59890184595581) (4.33209399070067,-2.97284896524537)};
\addplot [arrow1] coordinates{(4.21960104265042,-1.51263663360069) (4.56737994362305,-2.90669560360223)};
\addplot [arrow1] coordinates{(4.53146277194279,-1.44259701620221) (4.80580564476452,-2.8529468181788)};
\addplot [arrow1] coordinates{(4.84656269592522,-1.38897496748327) (5.04671758585095,-2.81174993065991)};
\addplot [arrow1] coordinates{(5.16403714760844,-1.35191746175217) (5.28954439974722,-2.7831785885068)};
\addplot [arrow1] coordinates{(5.48301595158766,-1.33152607105629) (5.53335389160737,-2.76742880717669)};
\addplot [arrow1] coordinates{(5.80262480913323,-1.32785668678014) (5.77774441998894,-2.76442605233111)};

\addplot [arrow2] coordinates{(-.570565567455198,-8.28765725056961) (-1.74133371699788,-8.08660883413348)};
\addplot [arrow2] coordinates{(-.518349601852232,-7.97228699362178) (-1.69168743120263,-7.78682307812778)};
\addplot [arrow2] coordinates{(-.470331956088577,-7.65625026924561) (-1.64603244434611,-7.4864065538891)};
\addplot [arrow2] coordinates{(-.426521465109313,-7.33960291605671) (-1.60437671439874,-7.1854097337936)};
\addplot [arrow2] coordinates{(-.388867251169302,-7.02217723210811) (-1.57129129905641,-6.90819523040005)};
\addplot [arrow2] coordinates{(-.365434653967573,-6.70340565056357) (-1.55213678998008,-6.6499587973501)};
\addplot [arrow2] coordinates{(-.358566207387946,-6.38384944148341) (-1.54643978571883,-6.39250320589304)};
\addplot [arrow2] coordinates{(-.368423495026279,-6.06437154989079) (-1.55421647872624,-6.1351778034496)};
\addplot [arrow2] coordinates{(-.394987041124122,-5.74584735240441) (-1.57544924795111,-5.87861596310924)};
\addplot [arrow2] coordinates{(-.438184429862013,-5.42914990717117) (-1.61008028390336,-5.62351717040182)};
\addplot [arrow2] coordinates{(-.497897280610775,-5.11514726122937) (-1.65801469565544,-5.37058044044689)};
\addplot [arrow2] coordinates{(-.573961925654153,-4.80470007413147) (-1.71912110041848,-5.12049904529536)};
\addplot [arrow2] coordinates{(-.666169877141864,-4.49865926004781) (-1.79323200992081,-4.87395844015731)};
\addplot [arrow2] coordinates{(-.774268399490989,-4.19786365557779) (-1.88014429131934,-4.6316343756684)};
\addplot [arrow2] coordinates{(-.897961202165748,-3.90313772056525) (-1.97961972409248,-4.39419104524431)};
\addplot [arrow2] coordinates{(-1.03690925179181,-3.61528927831193) (-2.09138565299347,-4.1622792645492)};
\addplot [arrow2] coordinates{(-1.19073170142328,-3.33510730138992) (-2.21513573537996,-3.93653468765394)};
\addplot [arrow2] coordinates{(-1.35900693441782,-3.06335974912278) (-2.35053078087789,-3.71757606475249)};
\addplot [arrow2] coordinates{(-1.54127372005845,-2.80091462662322) (-2.49719968107873,-3.50600354621061)};
\addplot [arrow2] coordinates{(-1.73703247775493,-2.54812212343082) (-2.65474042672188,-3.30239703759538)};

\node[aircraft top,draw=white,fill=black,minimum width=1cm,rotate=78,scale = 0.55] at (axis cs:-1.59531614,	-12.641604) {};
\node[aircraft top,draw=black,fill=mycolor2,minimum width=1cm,rotate=118,scale = 0.55] at (axis cs:15.8494239621087, -4.97744913675632) {};

\end{axis}
\end{tikzpicture}%
	\caption{\edit{Two s}imulated encounter trajectories}
	\label{fig_traj}
\end{figure}

\Cref{fig_traj} shows \edit{top-down views of} trajectories taken by aircraft in two example encounters, \edit{and although the advisories are horizontal, the vertical separation between the ownship and intruder aircraft is considered by the system through the $\tau$ parameter}. The intruder aircraft follows its own flight path while the ownship follows either the table or neural network collision avoidance system. The advisories given by the collision avoidance systems are represented by the arrows. In the first encounter, the neural network collision avoidance system alerts the ownship seven seconds earlier than the score table, which allows the aircraft to avoid a near mid-air collision. In the second encounter, the neural network chooses a different turning direction than the score table and avoids a near mid-air collision. As a continuous function, the neural network learns a better interpolation than the nearest-neighbor interpolation used by the original score table, allowing the network to outperform the original table in some cases.

\begin{figure}
	\input{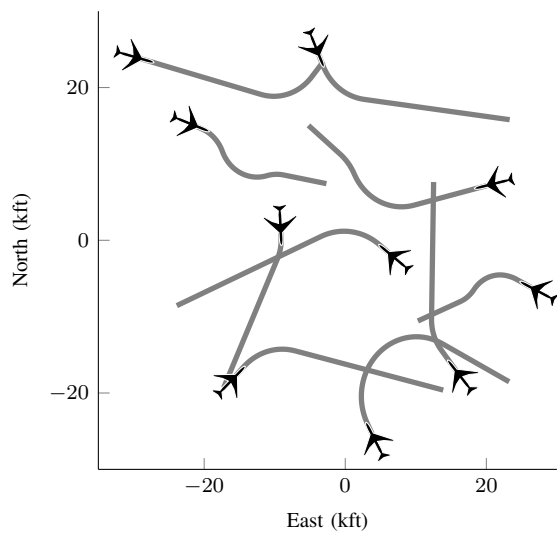}
	\caption{\edit{A simulated multi-intruder encounter trajectories}}
	\label{fig_trajMulti}
\end{figure}

\edit{
The neural network collision avoidance system is also effective in multi-intruder scenarios through utility fusion, which combines the individual decisions when each intruder is considered separately to generate an overall decision for the ownship \cite{Kochenderfer2011atc371}. \Cref{fig_trajMulti} shows an example scenario with ten aircraft all using the neural network collision avoidance policy. The aircraft positions and velocities are randomly initialized and commanded to fly straight unless the neural network alerts the aircraft to turn. The aircraft fly for 80 seconds, and all ten aircraft are safely routed around each other with a minimum separation of 5564 ft. A few aircraft experience reversals in their turning direction as collisions with different intruders become more imminent, illustrating that the neural network can safely maneuver aircraft through multi-intruder encounters.
}

While simulations can check the network performance in millions of states, simulations are not enough to guarantee the network will behave correctly in all possible states. One method for verifying the neural network representation is to use formal methods to prove properties about the network. The Reluplex algorithm represents neural networks with ReLU activations as a system of equations and uses a simplex method to find a set of inputs to satisfy a desired output constraint \cite{katz2017reluplex}. If no set of inputs can be found, then it is guaranteed that no set of inputs exist that satisfy the condition. This approach can be used to verify that the network will always alert if an intruder is nearby, for example.  \edit{The use of neural networks in safety-critical certified avionics is unprecedented, but} future work with Reluplex \edit{could provide a method for} verification and certification of neural networks in aircraft collision avoidance systems.

\section{Conclusions \label{sec:Conclusion}}
The \edit{unmanned variant of the Airborne Collision Avoidance System X (ACAS Xu)} makes decisions using an optimized score table, but the table is too large to be used in current avionics systems. A deep neural network representation was trained to approximate the table, maintaining optimal advisories while also approximating table values. Simulation shows that the compression algorithms perform as well as the original table. By factoring the table into subtables to train multiple small networks, the network's required runtime was reduced to the level of the original table lookups. By investigating areas of the state space where the networks represent the table poorly, the network training was modified to allow the network to train more easily, resulting in a more accurate representation of the original table. The neural network representation can be used in place of the original table without degrading performance, \edit{enabling} the collision avoidance system to be used in existing avionics hardware. \edit{If avionics systems with greater storage capacity become available, then this neural network compression approach will allow an even larger and more complex collision avoidance policy to be efficiently represented.}

\section*{Acknowledgments}
This work is sponsored by the Federal Aviation Administration under U.S. Air Force contract no. FA8721-05-C-002. The authors wish to thank Neal Suchy for his support. Opinions, interpretations, conclusions, and recommendations are
those of the authors and are not necessarily endorsed by the U.S. Government.

\bibliographystyle{IEEEtran}
\bibliography{references}
\end{document}